\documentclass[journal, onecolumn]{IEEEtran}
\ifCLASSINFOpdf
\else
\fi

\usepackage{bm}
\usepackage{CJK}
\usepackage{indentfirst}
\usepackage{amsmath}
\usepackage{multirow}
\usepackage{threeparttable}
\usepackage{graphicx}
\usepackage{subfigure}
\usepackage{color}
\usepackage{booktabs}
\usepackage{epstopdf}
\usepackage{makecell}
\usepackage[linesnumbered,ruled,vlined]{algorithm2e}
\usepackage{algpseudocode}
\usepackage{amsmath}
\usepackage{setspace}

\DeclareMathAlphabet{\mathcal}{OMS}{cmsy}{m}{n}

\begin{document}
\title{A Hybrid Deep Learning Model-based \\Remaining Useful Life Estimation for \\ Reed Relay with Degradation Pattern Clustering}

\author{Chinthaka~Gamanayake,~\IEEEmembership{}
        		Yan~Qin,~\IEEEmembership{Member,~IEEE,}
        		Chau~Yuen,~\IEEEmembership{Fellow,~IEEE,}
        		Lahiru~Jayasinghe,~\IEEEmembership{}
           		Dominique-Ea~Tan,~\IEEEmembership{}
           		and Jenny~Low\IEEEmembership{}
\thanks{Corresponding author: Yan Qin.}
\thanks{C. Gamanayake, Y. Qin, C. Yuen, and L. Jayasingheare are with the Engineering Product Development Pillar, The Singapore University of Technology and Design, 8 Somapah Road, 487372 Singapore. (e-mail: chinthaka\_madhushan@sutd.edu.sg, zdqinyan@gmail.com, yuenchau@sutd.edu.sg, lahiruaruna@gamil.com)}
\thanks{D. Tan and J. Low are with the Keysight Company, Singapore. (e-mail: dominique-ea$\_$tan@keysight.com, jenny-cn$\_$lowg@keysight.com)}
}

\maketitle
\begin{abstract}
Reed relay serves as the fundamental component of functional testing, which closely relates to the successful quality inspection of electronics. To provide accurate remaining useful life (RUL) estimation for reed relay, a hybrid deep learning network with degradation pattern clustering is proposed based on the following three considerations. First, multiple degradation behaviors are observed for reed relay, and hence a dynamic time wrapping-based $K$-means clustering is offered to distinguish degradation patterns from each other. Second, although proper selections of features are of great significance, few studies are available to guide the selection. The proposed method recommends operational rules for easy implementation purposes. Third, a neural network for remaining useful life estimation (RULNet) is proposed to address the weakness of the convolutional neural network (CNN) in capturing temporal information of sequential data, which incorporates temporal correlation ability after high-level feature representation of convolutional operation. In this way, three variants of RULNet are constructed with health indicators, features with self-organizing map, or features with curve fitting. Ultimately, the proposed hybrid model is compared with the typical baseline models, including CNN and long short-term memory network (LSTM), through a practical reed relay dataset with two distinct degradation manners. The results from both degradation cases demonstrate that the proposed method outperforms CNN and LSTM regarding the index root mean squared error.
\end{abstract}


\IEEEpeerreviewmaketitle

\section{Introduction}
\IEEEPARstart{T}{he} rising deployment of digital solutions attracts extensive attention ranging from various companies to governments, meeting changes in an era of digital transformation \cite{Heatpump}. This promising transformation requires the strong support of high-quality printed circuit boards (PCBs) products. That is, by connecting each functional electronics component together, PCBs enable end-users to communicate, monitor, and control connected devices and systems wirelessly in real-time. To avoid unqualified PCBs flowing into the market, strict testing with thousands of test points is indispensable. Reed relay has been widely adopted in functional testing of PCBs for switching control purposes, ascribing to the benefits of high electrical off-isolation, low on-resistance, and the excellent ability to withstand electrostatic discharge \cite{Reed_Relay}. Each test point will need frequent switches among different voltages and currents to finally determine functionality \cite{PCB_background}. In practice, the natural aging of the reed relay is inevitable, and this will cause significant maintenance expenditures or production downtime if the failed reed relays could not be timely replaced. Therefore, accurate remaining useful life (RUL) estimation makes it possible to schedule the replacement of reed relays in advance, providing a reliable and safe operation environment for high-performance quality inspection of PCBs.

The emerging revolutions in information technology and artificial intelligence techniques have driven the researches on RUL estimation from model-based mechanistic analysis towards data-driven approaches with the assist of big data and ever-increasing computing ability \cite{CAA_Reciew2}. Although model-based methods excel in offering insights into failure mechanisms \cite{2021TCSTYin}, \cite{2022TBYin}, the requirement of a substantial amount of domain knowledge may significantly weaken its advantages, especially for complex devices. Alternatively, in the era of big data and artificial intelligence, data-driven RUL models are capable of evaluating the health status of crucial assets with affordable costs, detecting abnormal behaviors, and estimating RUL in advance. In this way, much manpower can be saved from complex degradation mechanism analysis in light of the deep presentation capability of machine learning methods.

\begin{figure}[!htb]
\subfigure[]
{
\begin{minipage}[t]{1\linewidth}
\centering
\includegraphics[width=6cm]{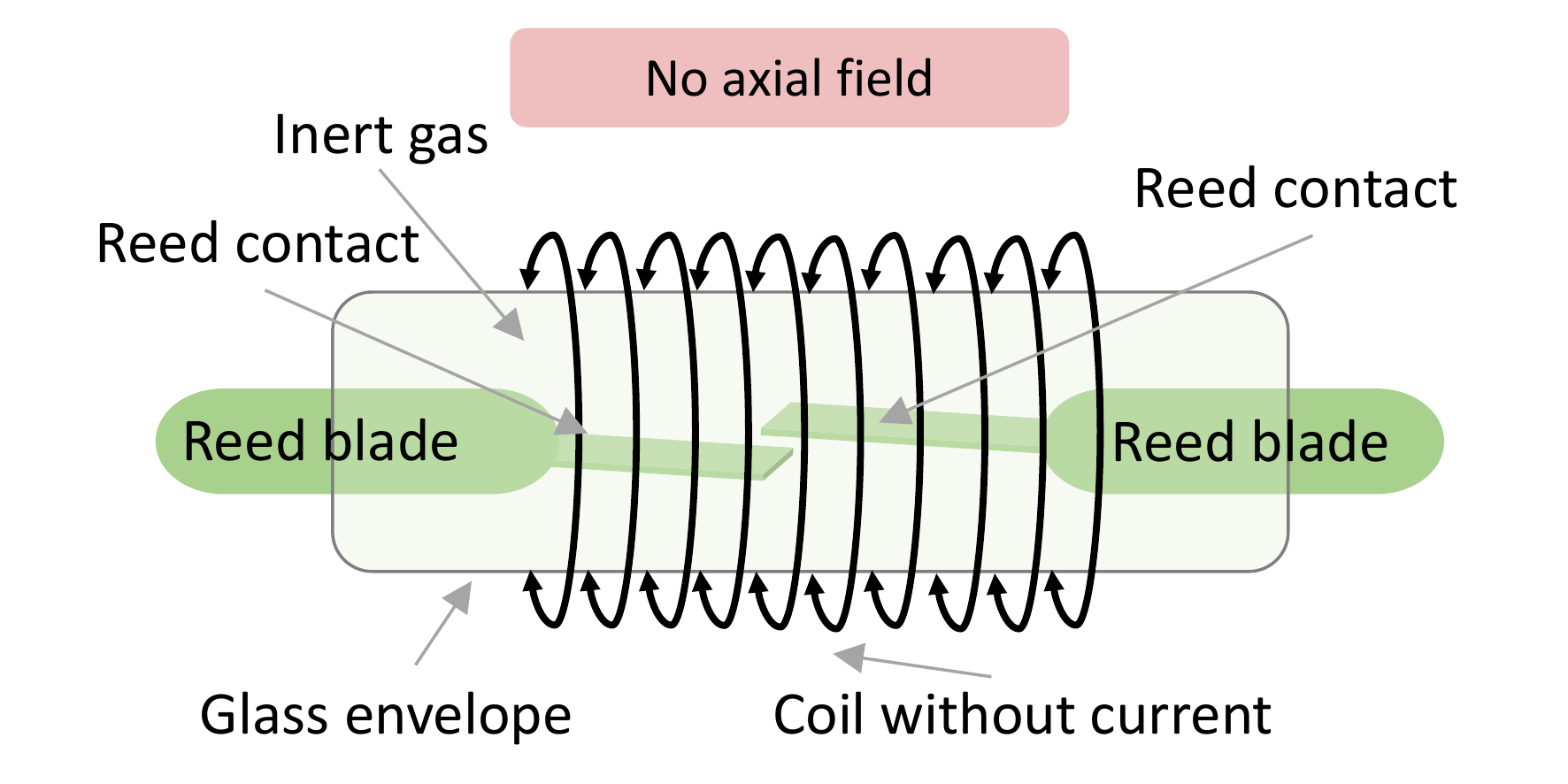}
\end{minipage}
}
\hfill
\subfigure[]
{
\begin{minipage}[t]{1\linewidth}
\centering
\includegraphics[width=6cm]{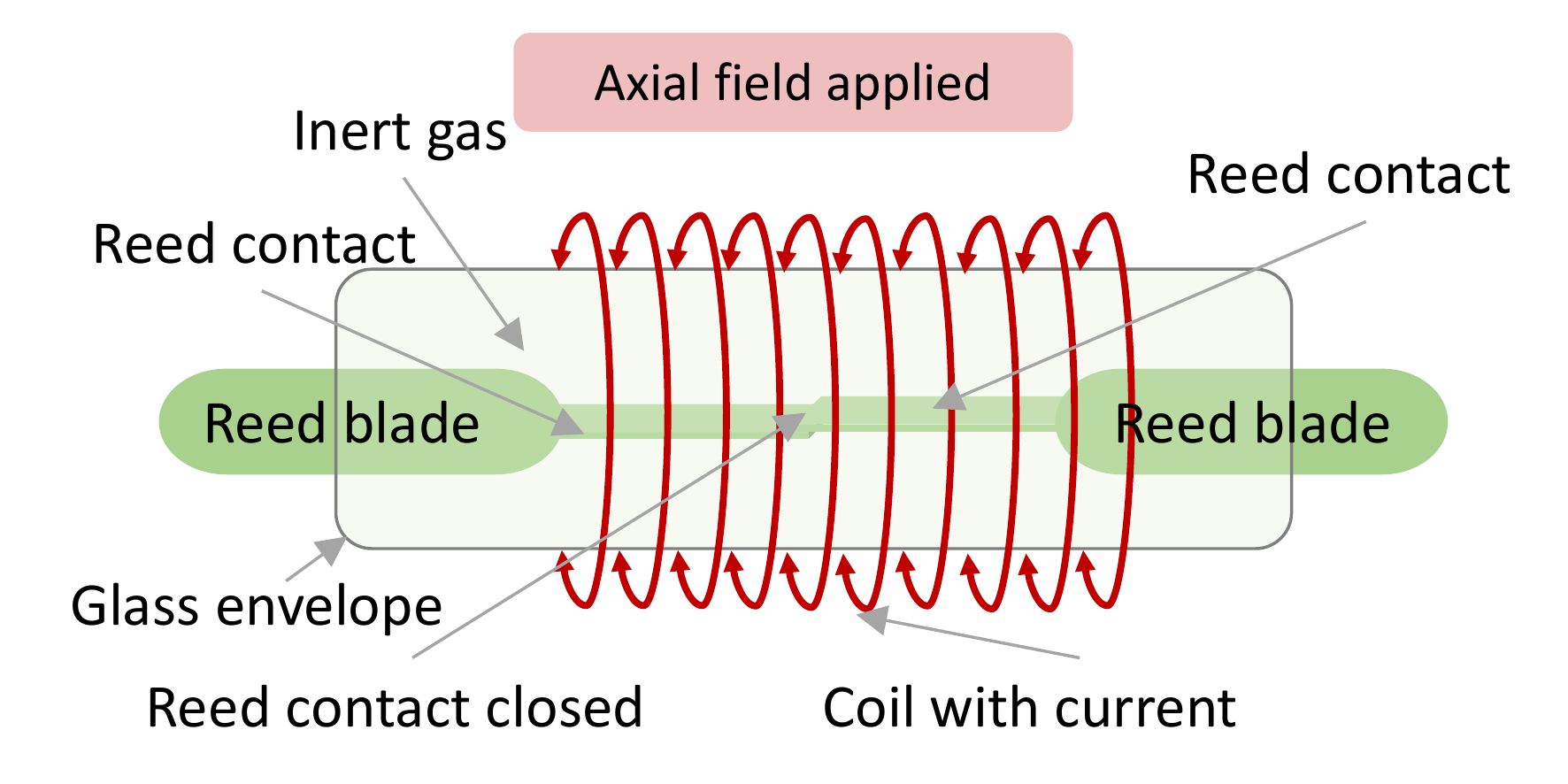}
\end{minipage}
}
\caption{Illustration of (a) The basic elements of reed relay and (b) the working mode of reed relay when axial field is applied.}
\label{FIGTII1}
\end{figure}

The high safety requirements of critical assets have sparked a flurry of interests in the research area of data-driven RUL estimation. Fruitful works regarding RUL have been observed in typical scenarios, such as rolling bearing \cite{BearingRUL}, \cite{BearingRUL1}, aircraft engine system \cite{GANRUL}, \cite{TemCaps}. The rolling bearing, which is the central part of the rotating machinery system, is often assessed by the one-dimensional vibration signal with a high sample frequency. On the other hand, the aircraft engine has several components and produces numerous time series for health inference. Lei et al. \cite{Lei2018Review} demonstrated the research trend from four key perspectives in the comprehensive survey, encompassing data collection, the creation of health indicators, the division of health stages, and the RUL prediction model applying statistical machine learning approaches.

Deep learning technologies are now receiving increasing attention in RUL research field, as opposed to shallow machine learning techniques. Deutsch et al. \cite{DBNetwork} proposed a RUL estimation model with deep belief network, which has the benefits of powerful feature representation, with the goal of extracting deep features from bearing vibration data. Babu et al. \cite{Babu2016CNN_RUL} proposed the RUL estimation model with two-convolutional layers for a multivariate engine system. In this work, each piece of complete run-to-failure time-series was sliced into a series of two-dimensional data slices with a fixed window length and fed into the convolutional neural network (CNN) model in the image format \cite{CNN2010}. To combat over-fitting, Li et al. \cite{CNN_2018} incorporated dropout technology into a deep CNN-based RUL estimation model. Liu et al. \cite{JL_CNN2020} proposed a novel CNN network by optimizing the loss jointly constructed from the classification and prediction tasks, simultaneously considering both fault diagnosis and prognostic in a unified framework. To address the data scarcity issue of run-to-failure degradation procedure, Zhang et al. \cite{GANRUL} attempted to improve estimation performances by using more realistic-like time-series generated by the newly designed generative adversarial network, while keeping the model structure of the previous RUL estimation models the same. Apart from CNN, another outstanding architecture of deep learning architecture is the recurrent neural network (RNN) \cite{RNN}, \cite{LSTMBattery}, which is explicitly designed explicitly for sequential learning. In light of the temporal correlations of time-series, Zheng et al. \cite{zheng2017LSTM_RUL} successfully applied the long short-term memory (LSTM) model, an imperative variant of RNN, for RUL estimation of the aircraft engine system. Recently, attention has been paid to emerging advanced models such as the attention mechanism \cite{2020WangTIE} and graph neural network \cite{2022GNNTIM}, \cite{2021TAIGAN}, \cite{2022GNNReview}.

Critical assets are prone to follow several failure modes or working conditions caused by the interactive influences between the external environment and the internal dynamics. Taking advantage of the auxiliary variables, such as wind speed and ambient temperature, Rezamand et al. \cite{2021Rezamand} identified different failure dynamics from vibration signals. With experimental data of the multivariate aircraft engineering system, Huang et al. \cite{Mult_2018TIE} proposed a dual bidirectional LSTM-based feature extractor to handle the multiple working conditions. The second bidirectional LSTM network captures the variable working conditions, which is combined with the low-level features from the first bidirectional LSTM with measurements. Wu et al. \cite{WuTIM2021} put forward a degradation-aware RUL model for bearing data, assuming that the classification of historical observations from components/machines is known in the same operational environments as the testing samples. Recently, transfer learning has played an important role in handling new working conditions. Mao et al. \cite{2019Mao} separated the common fault features between the source data and the target bearing data used for the prediction with the least-square support vector machine. To transfer the rich knowledge from source data to the incomplete target data, Siahpour et al. \cite{2022TIM} achieved transfer learning with the proposed consistency-based regularization on the convolutional operation. He et al. \cite{2023He} reached a transferable neural network by simultaneously minimizing the distance between both marginal and conditional probability distributions in different domains. It is worth noting that the above-mentioned approaches necessitate \textit{a priori} knowledge or signals to indicate the multiple working conditions.

\begin{figure*}[!htb]
\subfigure[The first operating condition]
{
\begin{minipage}[t]{0.3\linewidth}
\centering
\includegraphics[width=6cm]{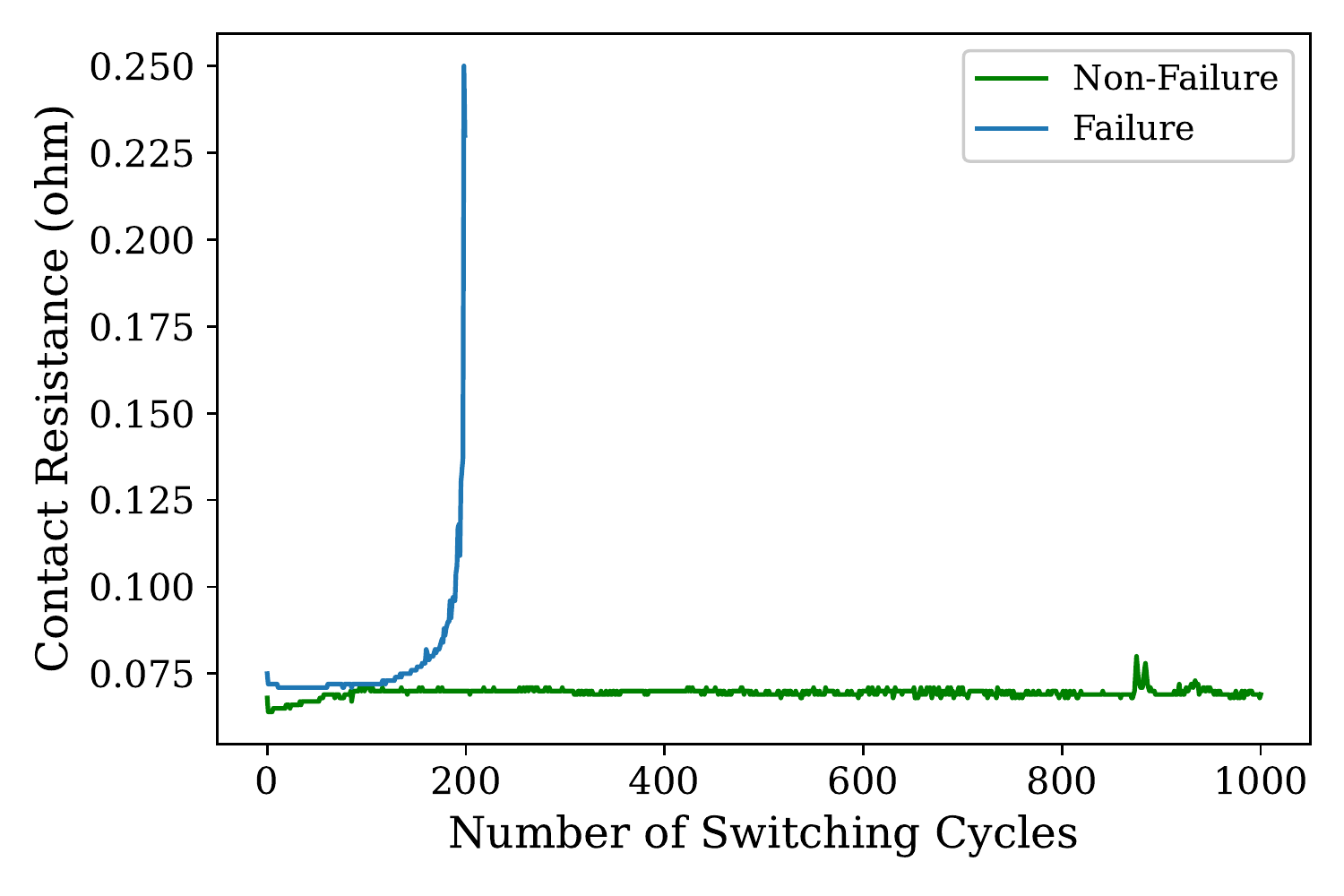}
\end{minipage}
}
\subfigure[The second operating condition]
{
\begin{minipage}[t]{0.3\linewidth}
\centering
\includegraphics[width=6cm]{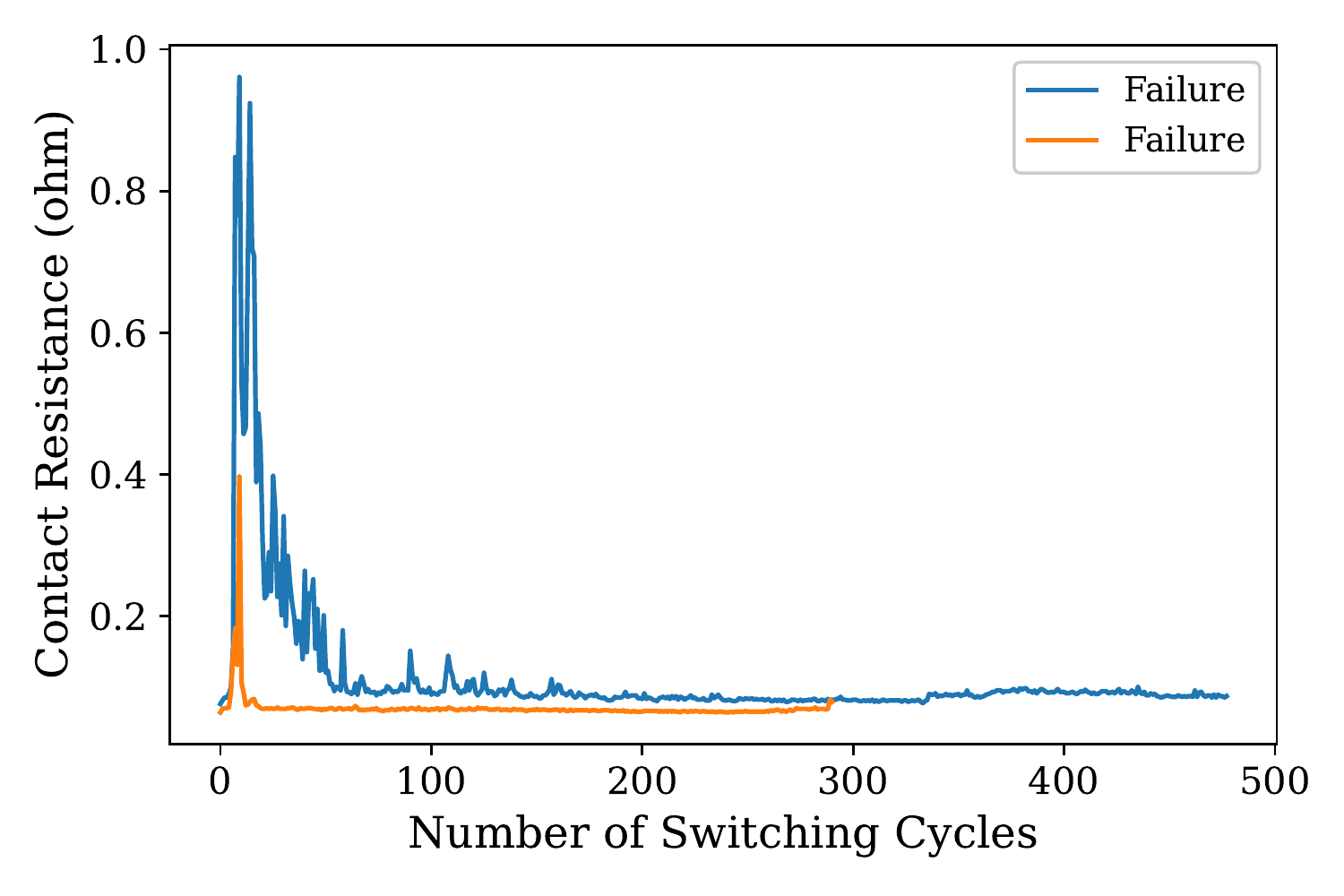}
\end{minipage}
}
\subfigure[The third operating condition]
{
\begin{minipage}[t]{0.3\linewidth}
\centering
\includegraphics[width=6cm]{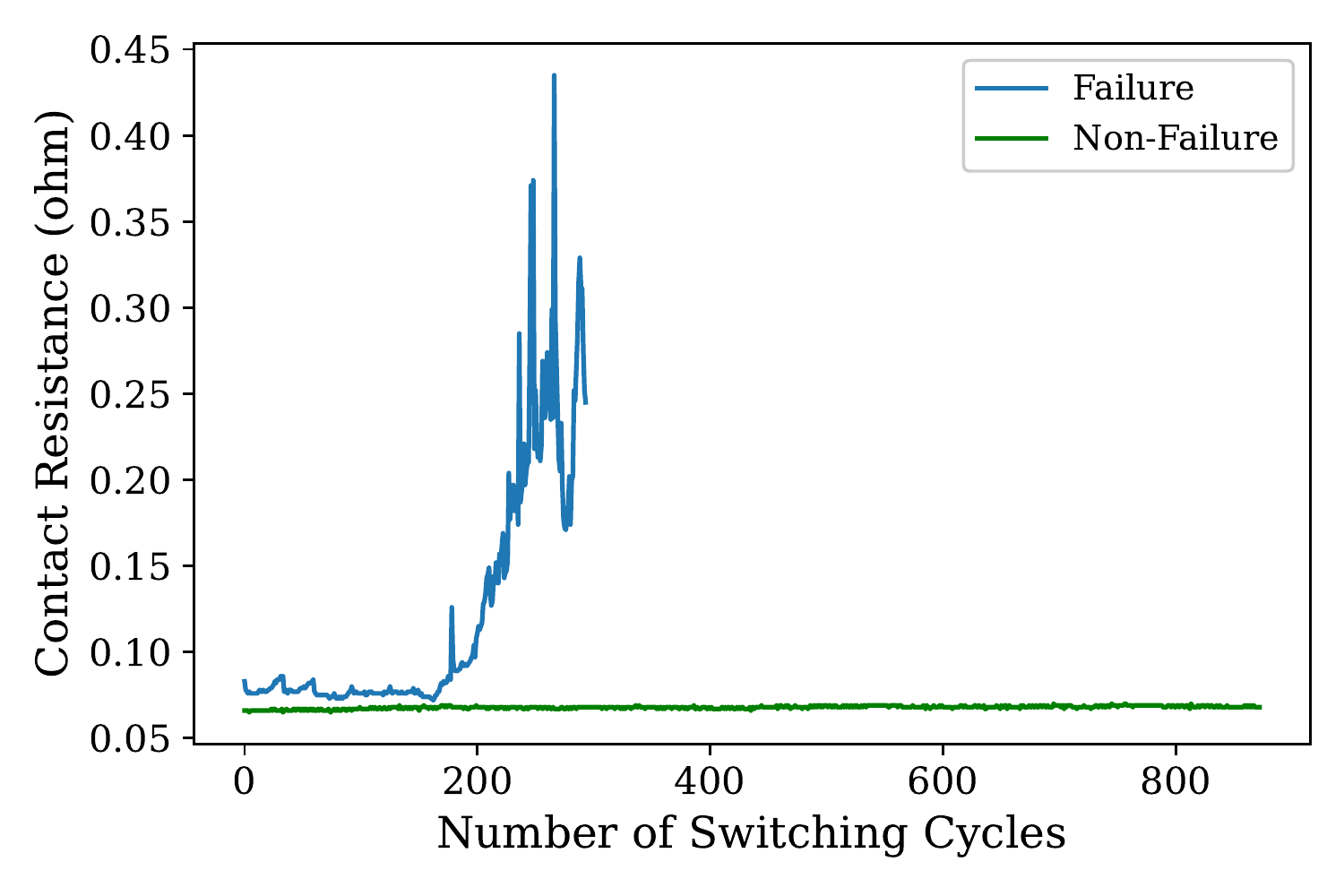}
\end{minipage}
}

\subfigure[The fourth operating condition]
{
\begin{minipage}[t]{0.3\linewidth}
\centering
\includegraphics[width=6cm]{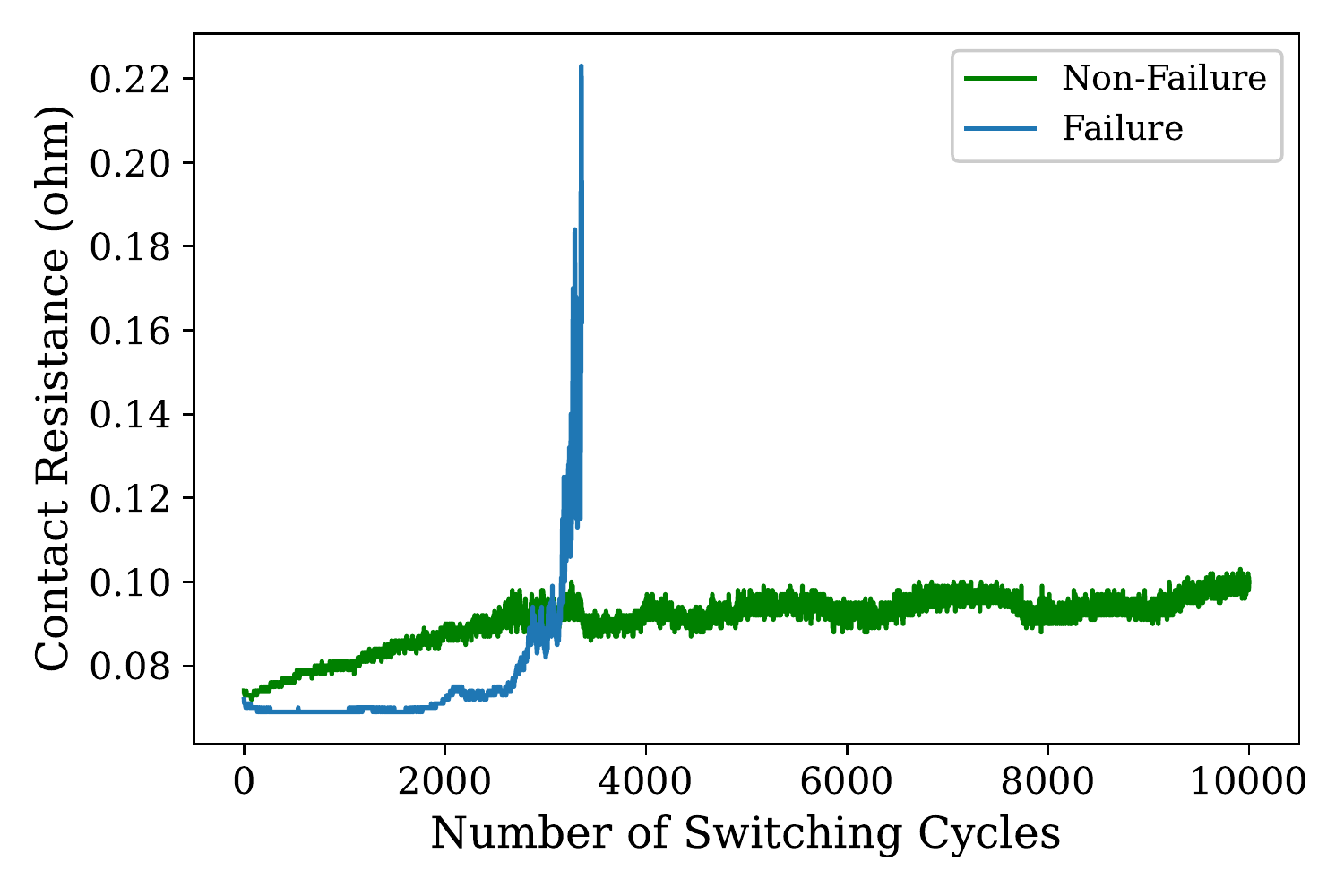}
\end{minipage}
}
\subfigure[The fifth operating condition]
{
\begin{minipage}[t]{0.3\linewidth}
\centering
\includegraphics[width=6cm]{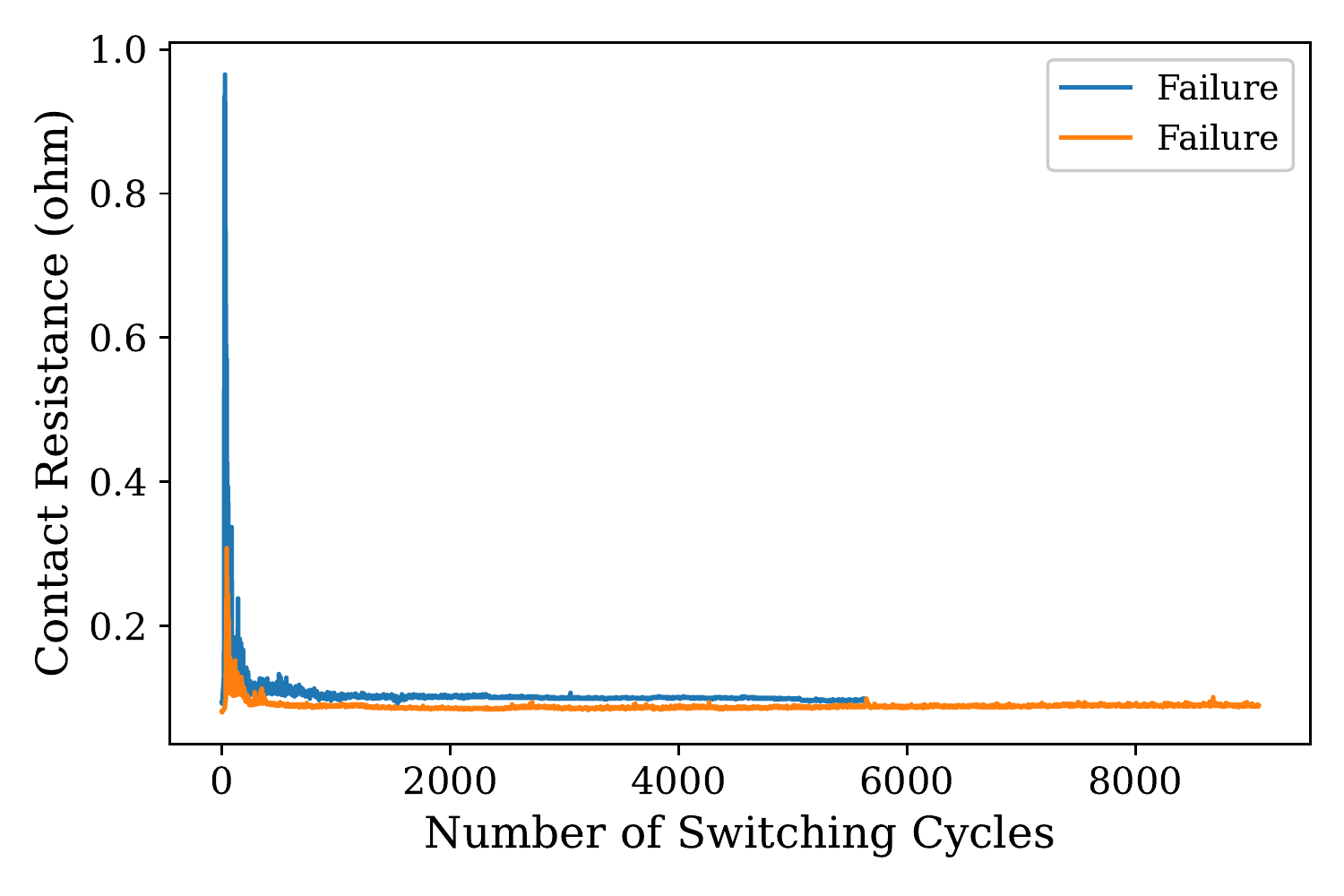}
\end{minipage}
}
\subfigure[The sixth operating condition]
{
\begin{minipage}[t]{0.3\linewidth}
\centering
\includegraphics[width=6cm]{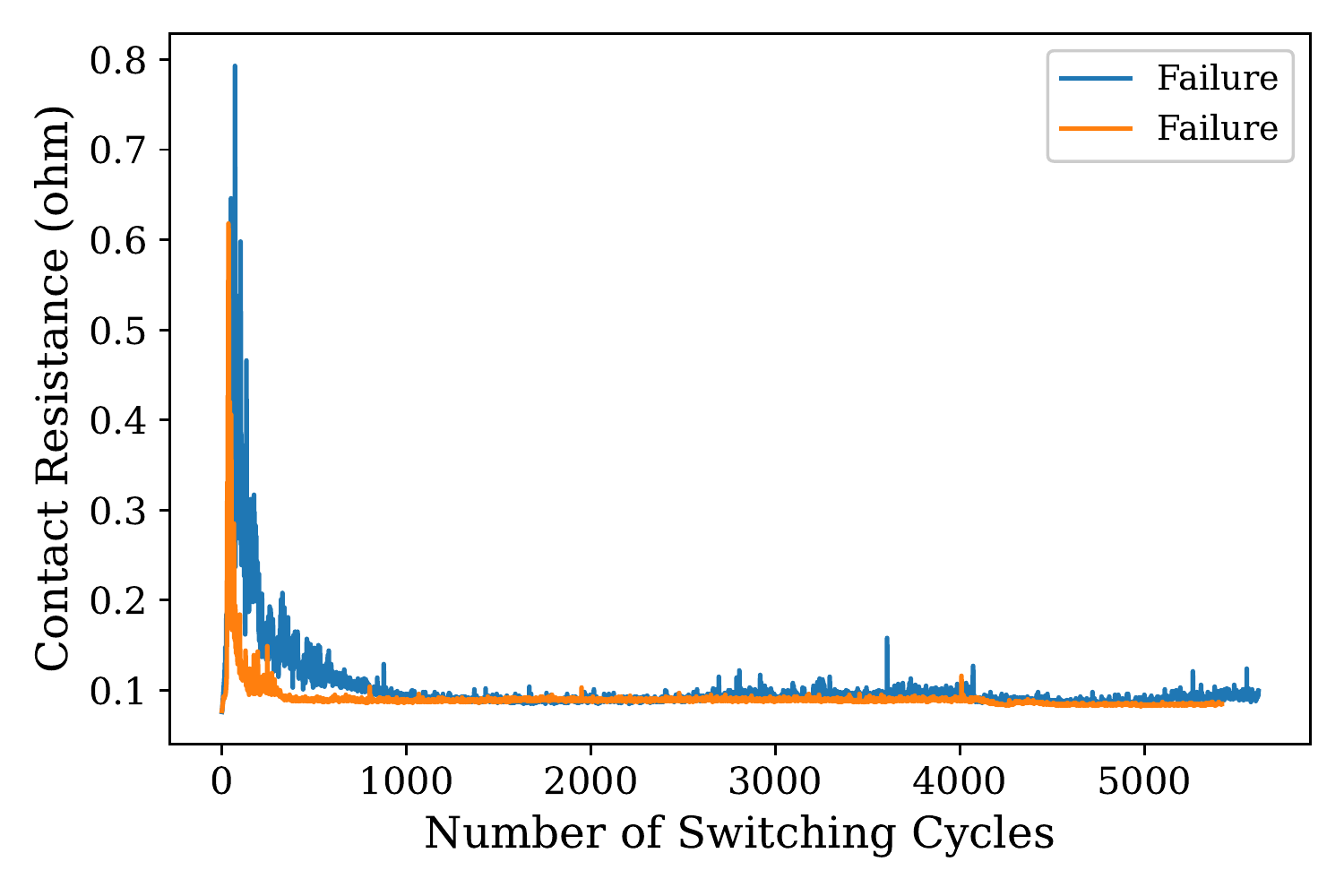}
\end{minipage}
}
\caption{The demonstration of raw data collected in the first six operating conditions.}
\label{FIGTII2}
\end{figure*}

Although it is the superiority of deep learning models to automatically extract in-depth representation from raw measurements, the inherent mechanistic dynamics may not be well represented. This weakness can be largely alleviated by hybrid models, which combine accessible domain knowledge with data-driven modeling approaches to bridge the gap between idealized mechanistic models and reality. Chen et al. \cite{AttenLSTM_2020} addressed the importance of feature engineering, and the processed features conveying domain knowledge contribute to improve RUL estimation performance. In \cite{MuScale2019}, the frequency information of bearing vibration signal was described using Short-time Fourier transform. Following that, a CNN model was used to perform the multi-scale feature extraction. To thoroughly decouple different frequencies, the Hilbert-Huang transform was adopted by Cheng et al. \cite{2021Cheng} for raw vibration measurement processing, which served as the input of deep CNN. Similarly, Wu et al. \cite{LSTM1_2018} utilized differential features of raw measurements into the LSTM-based RUL estimation model. These hybrid models [27]-[29] share the similarity that accessible domain knowledge guides the construction of features before they are fed into deep neural networks. This benefits the reduction of the estimation error for hybrid models compared to pure data-driven models. Although the above-mentioned approaches have been successfully reported in the RUL estimation issue, they still suffer from the following challenges concerning the unique degradation characteristics of reed relay:
\begin{itemize}
\item Degradation behavior in current literature is always assumed to follow a single pattern, which may not meet with the actual case since the reed relay experiences more than one kind of degradation pattern caused by the complex electrochemical characteristics. As such, identifying and clustering different patterns is still lacking.
\item Extracting the most relevant information to degradation procedure, feature engineering plays a fundamental but essential role in hybrid models. The quality of feature engineering has a significant influence on the final estimation ability. However, it is still not clear how to systematically summarize and compare typical features. Hence the construction of proper features for specific systems deserves further exploration.
\item CNN is capable of capturing in-depth features from image-format data, and LSTM goods at describing the temporal correlation among data time-series. However, it is still lacking to integrate the merits of both approaches to boost the RUL estimation performance.
\end{itemize}

\begin{table*}[!htb]
    \centering
	\caption{The brief summary of operating conditions in the employed testing machine.}
	\renewcommand{\arraystretch}{1}
    \scriptsize
	\begin{tabular}{p{1.2cm} p{2cm} p{2cm} p{2cm} p{2cm} p{2cm} p{2cm}}
		\midrule
        \midrule
		\textbf{No.} & \textbf{Drive frequency} & \textbf{Switch voltage} & \textbf{Switch current} & \textbf{Load resistance} & \textbf{Load wattage} & \textbf{Total million}\\
		\midrule
		{Condition 1} & 400 & 1 & 10 & 100 & 0.01 & 1000 \\
               \midrule
		{Condition 2} & 400 & 1 & 10 & 100 & 0.01 & 1000 \\
		\midrule
		{Condition 3} & 400 & 5 & 100 & 50 & 0.5 & 1000 \\
		\midrule
		{Condition 4} & 400 & 20 & 500 & 40 & 10 & 250 \\
		\midrule
		{Condition 5} & 400 & 12 & 4 & 3000 & 0.048 & 250 \\
		\midrule
		{Condition 6} & 400 & 5 & 10 & 500 & 0.05 & 1000 \\
		\midrule
        \midrule
\end{tabular}
\label{TableTII1}
\end{table*}

To address the concerned challenges, we propose a degradation pattern-based hybrid deep learning network for RUL estimation. The proposed approach starts by investigating the degradation behavior of the reed relay, and then degradation patterns are clustered using dynamic time warping (DTW) with $K$-means clustering algorithm. Next, three kinds of typical features are carefully constructed according to the demands of domain knowledge, which are health indicators, features with self-organization map (SOM), and features with curve fitting, respectively. Moreover, operational guidance is recommended to select proper feature engineering according to the data characteristics and accessible domain knowledge. Next, the network structure of the hybrid model is achieved through the proposed RULNet, which is a comprehensive combination of CNN and LSTM. Accordingly, with data collected from practical scenarios, three different hybrid models are developed with joint usage of the above-defined features and RULNet. Finally, performances of three different hybrid models are illustrated and further compared with traditional CNN and LSTM based hybrid models on an industrial reed relay dataset. For clarity, the main contributions of this work are summarized as follows:

\begin{itemize}
\item Failure patterns are carefully clustered and analyzed using DTW with $K$-means clustering algorithms, providing in-depth process understanding and facilitating the modeling for each pattern;
\item Three typical features, obtained from health indicator, SOM, and curve fitting, are carefully constructed and compared, delivering guidance for how to select a proper feature engineering;
\item An enhanced RUL estimation model is designed using the comprehensive integration of temporal correlation ability and convolutional feature representation.
\end{itemize}

The remainder of this article is structured as follows: Section II describes the collected reed relay data. Details of the hybrid estimation model for reed relay are given in Section III. Finally, we compare the prediction results of the different models on an industrial reed relay dataset in Section IV and conclude the article with a summarization in Section V.

\section{Process illustration and data description}
In this section, we describe the primary working mechanism of reed relays and collect a large amount of run-to-failure time-series to serve as the basis for our following analysis.

\subsection{The Working Mechanism of Reed Relays}
Fig. \ref{FIGTII1}(a) shows the typical components of a reed relay, consisting of two reed blades, two reed contacts, a coil, and a glass envelope with inert gas. As external current is passed through the coil, an axial magnetic field is created, and the reed contacts made of ferromagnetic material become magnetized. With the increasing magnetic field, the open reed contacts are attracted to each other, and the blades deflect to close the gap, as shown in Fig. \ref{FIGTII1}(b). On the contrary, removing the current will release the field, and the contacts then spring apart. Glass envelope not only holds the reed blades in place but also provides a hermetic seal to prevent any contaminants from entering the critical contact areas.

\subsection{Data Description}
The reed relay data used in our experiment are provided by Keysight Technologies in Singapore. In the dataset, the relays are switched on and off millions of cycles until the relay fails or a specific limit during the accelerated lifetime test. Due to the various requirements under testing, a total number of 21 operating conditions composed of different switch types, switch voltages, switch currents, etc., are observed. A batch of testing reed relays will be randomly assigned to cover all operating conditions. One reed relay under a specific operating condition produces a univariate time series containing a series of positive resistance values. Typically, contact resistance is defined as the electrical resistance of a reed contact in the closed state \cite{2021NatureCR}. During the accelerated lifetime test, the contact resistance of the reed relays is measured and logged. Eventually, time-series data from the recorded contact resistance are used as the raw data to predict RUL of reed relays. For better understanding, Table \ref{TableTII1} summarizes the detailed information of the first six operating conditions. In each operating condition, raw normal time series and several failure time series are demonstrated in Fig. \ref{FIGTII2}. It is observed that the contact resistance of normal time-series slightly varies within a low threshold throughout its lifespan. Contrarily, time series of failure reed relays follow a quickly changed behavior, and multiple degradation patterns exist due to the interactive influences of the unexpected faults.

During the accelerated lifetime test, the relay may suffer from the following unexpected fault at a certain time cycle:
\begin{itemize}
\item Sticking: reed relays do not open when they should at an individual time;
\item Missing: reed relays fail to close when they should at an individual time;
\item Drifting: static contact resistance gradually drifts up to an unacceptable level in a continuous time period.
\end{itemize}

Some reed relays finally reach the failure condition with the negative influences of continuous faults. However, due to various operating conditions, observation of the multiple failure patterns attracts our attention, and further analysis of their specific behavior needs additional effort. Here, the total number of 248 time series reaches the failure state for RUL modeling purposes.

\begin{figure*}[!htb]
\centering
\includegraphics[scale=0.25]{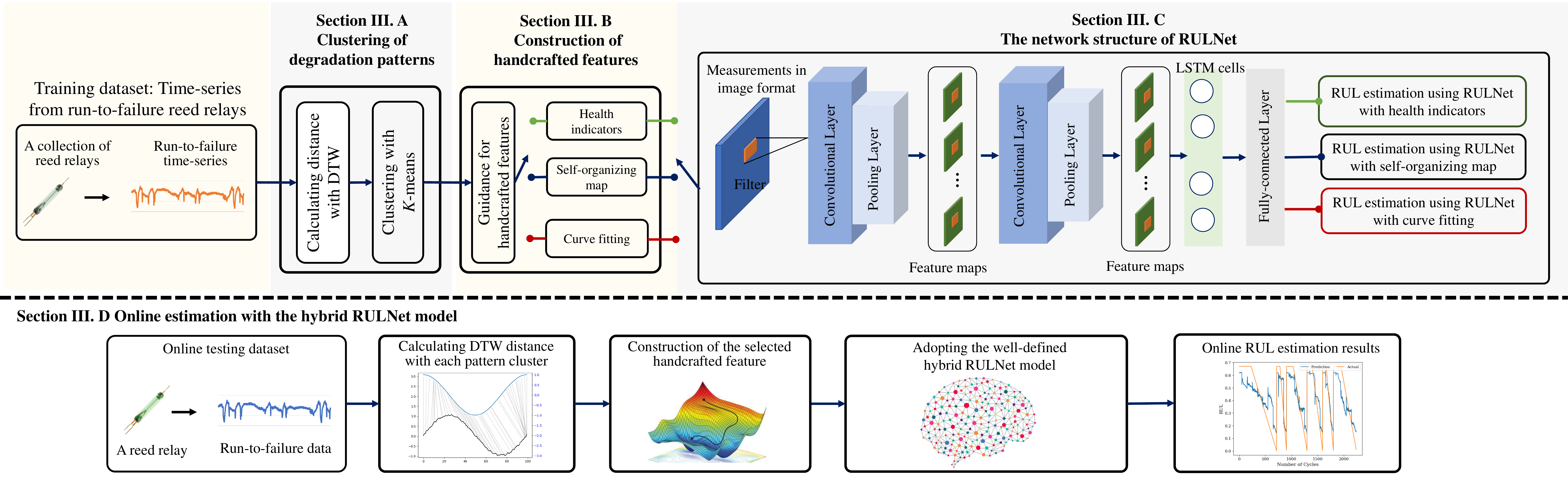}\\
\caption{The overall structure of the proposed hybrid estimation model for reed relay.}
\label{FIGTII3}
\end{figure*}

\begin{algorithm}[!htb]
\caption{The training steps of SOM}
\scriptsize
\LinesNumbered
\KwIn{Normalized time-series $\mathbf X_m$ and SOM model with $U$ map units}
\KwOut{A well-trained SOM network when the iteration epoch reaches its threshold or no further performance improvement is observed.}
Initialize weights of each node in SOM randomly\\
Give the value of iteration epoch $k$ to be 1 \\
\For {$n = 1, 2, \cdots, N$}
{
{{Selecting the $n$-th sample $\mathbf x_n$ from $\mathbf X_m$}}\\
	\For {$u= 1, 2, \cdots, U$}
	{
	\quad {Calculating the the Euclidian distance $D_{n, u}$ between $\mathbf x_n$ and the weights of the $u$-th map unit} \\
	}
	\quad {Identifying the best matching unit (BMU) with the minimum among $D_{n, u}$;} \\
	\quad {Updating the weights of BMU according to $\mathbf u_{BMU}(k+1)=\mathbf u_{BMU}(k) + \alpha(k)(\mathbf x_n - \mathbf u_{BMU}(k))$, where $0<\alpha(k)<1$ is the learning rate, and it is a decreasing function;} \\
	\For {$i= 1, 2, \cdots, U-1$}
	{
	\quad {Updating the weights of the $i$-th neighbors of BMU according to $\mathbf u_i(k+1)=\mathbf u_i(k) + \alpha(t)H(\mathbf u_{BMU}, i, k)(\mathbf x_n - \mathbf u_i(k))$, where $H(\mathbf u_{BMU}, i, k)$ is the neighborhood function regarding the distance between the BMU and $i$-th unit in the network, and the iteration time $k$;} \\
	}
	\quad {Updating $n=n+1$ and $k=k+1$} \\
}
\textbf{end for}
\end{algorithm}

\section {Hybrid RUL Estimation Model with Feature Engineering and RULNet}
With degradation pattern clustering and identification, the proposed hybrid estimation model consists of two sequential parts, as shown in Fig. \ref{FIGTII3}, i.e., the offline modeling stage and the online estimation stage. In the offline modeling stage, the proposed hybrid approach consists of three sequential steps. In the online estimation stage, when a new reed relay time-series is being tested, its belonging to the specific degradation pattern will be identified first, and then the well-selected RUL estimation model can be adopted for RUL estimation purpose.

\subsection{Clustering of Degradation Patterns}
Clustering time-series is crucial before applying RUL calculation methodologies since different degradation patterns need to be treated separately to achieve higher accuracy in RUL estimation. Usually, time-series data are compared with each other to cluster the same patterns. The objective of a time-series comparison method is to produce a distance metric between two input time-series. The similarity or dissimilarity of two-time series is typically calculated by converting the data into vectors and calculating the Euclidean distance between those points in vector space. If two time-series are highly correlated, but one is shifted by even a one-time step, Euclidean distance would erroneously measure them as further apart. Instead, it is better to use DTW to compare time-series. DTW is one of the algorithms for measuring the similarity between two temporal time-series sequences that do not align exactly in time \cite{petitjean2011global}.

Given time-series $\mathbf x$ and time-series $\mathbf y$, the DTW distance from $\mathbf x$ to $\mathbf y$ is calculated as the squared root of the sum of squared distances between each element in $\mathbf x$ and its nearest point in $\mathbf y$. DTW can be formulated as the following optimization problem below,
\begin{equation}\label{eq:DTW}
D(\mathbf x, \mathbf y) = min \sqrt{\sum_{(i,j) \in \pi} d(x_i, y_j)^2}
\end{equation}
where $\pi$ stands for a path between $\mathbf x$ and $\mathbf y$; $d(\cdot)$ measures the Euclidean distance between $x_i$ and $y_j$, in which $x_i$ and $y_j$ are the $i^{th}$ and the $j^{th}$ sampling in $\mathbf x$ and $\mathbf y$, respectively.

Instead of calculating Euclidean distance between corresponding elements from $\mathbf x$ and $\mathbf y$, DTW distance uses the element from $\mathbf y$, which is the nearest from that of $\mathbf x$. Therefore, DTW distance is more representative than Euclidean distance for pair-wise time-series comparison.

\begin{figure}[!ht]
\centering
{\includegraphics[width=0.35\textwidth]{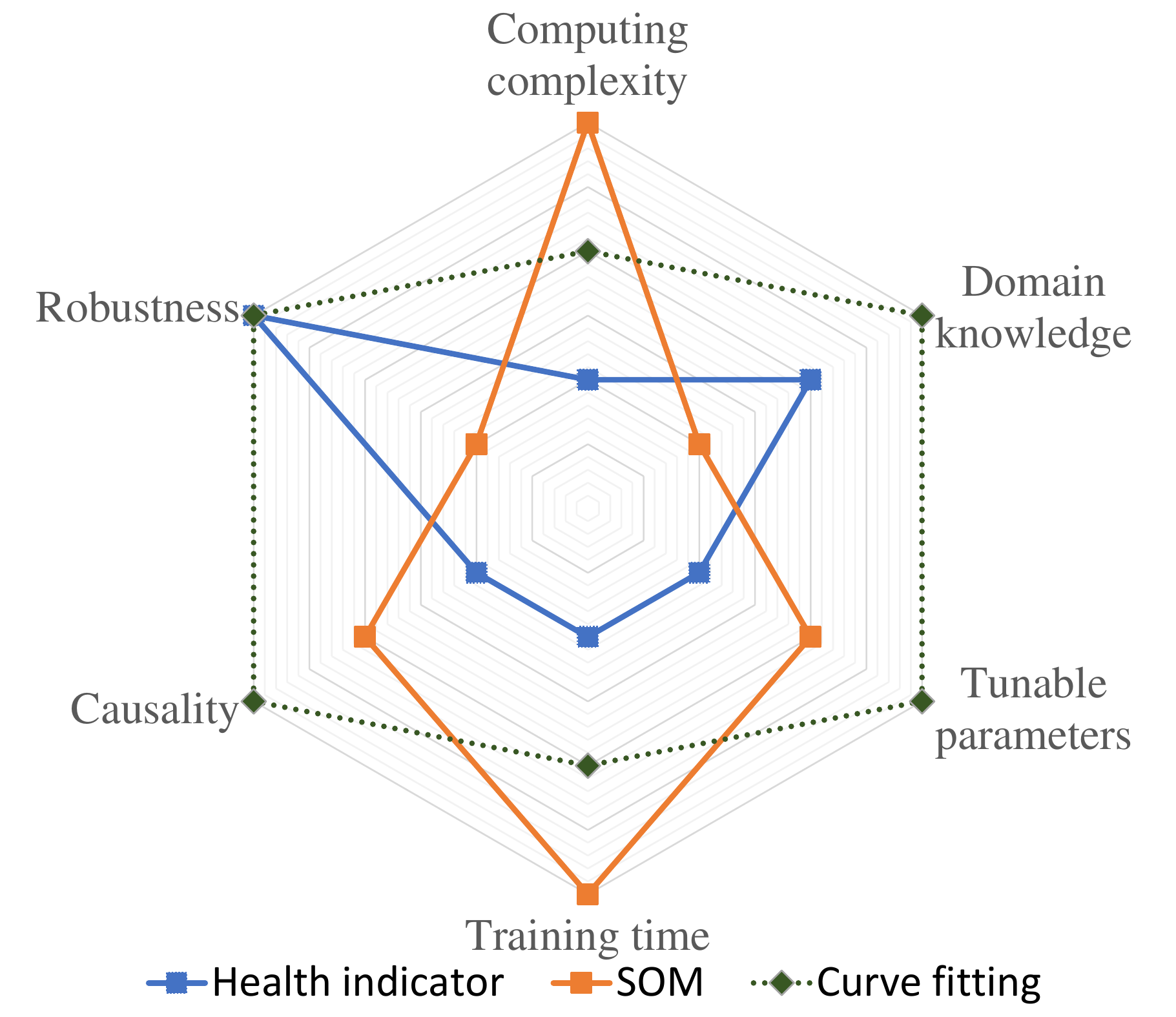}}
\caption{Comprehensive comparisons of the proposed feature engineering ways in six aspects.}
\label{FIGTII4}
\end{figure}

\begin{figure}[!ht]
\centering
{\includegraphics[width=0.45\textwidth]{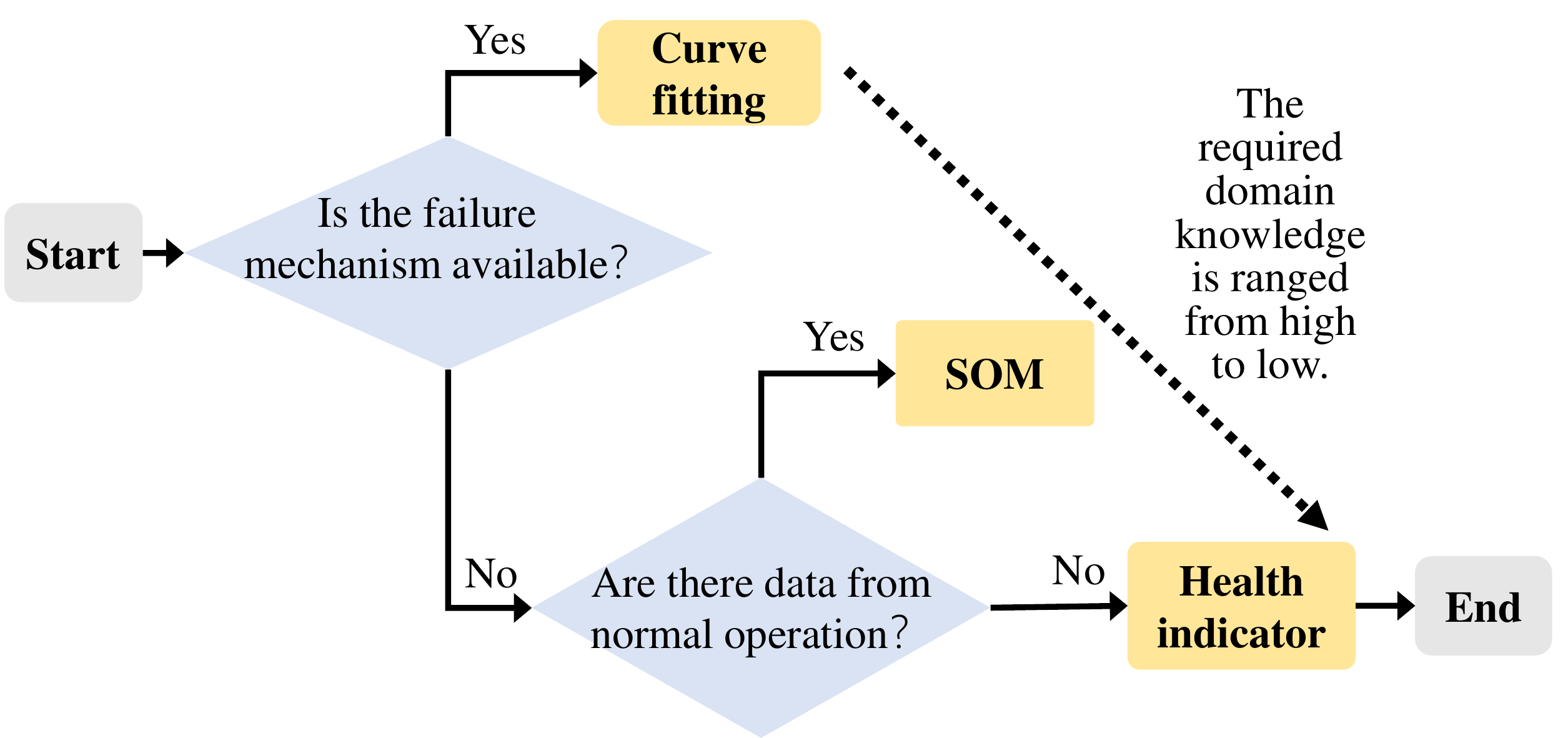}}
\caption{A general guidance for the selection of a proper feature.}
\label{FIGTII5}
\end{figure}

Assuming that $C$ run-to-failure time-series are available, distance matrixes with DTW could be calculated by performing Eq. (1) for any two time-series, which is given below,
\begin{equation}
\begin{aligned}
\mathbf{D}
&=\left\{\begin{array}{llllll}
D_{1,1} & D_{1,2} & \cdots & D_{1,c} & \cdots & D_{1,C} \\
D_{2,1} & D_{2,2} & \cdots & D_{2,c} & \cdots & D_{2,C} \\
\cdots   & \cdots  & \cdots & \cdots & \cdots & \cdots \\
D_{C,1} & D_{C,2} & \cdots & D_{C,c} & \cdots & D_{C,C} \\
\end{array}\right\}
\end{aligned}
\end{equation}
where $D_{i,j} (i,j \in[1, C])$ calculates the distance between the $i^{th}$ time-series and the $j^{th}$ time-series. It is easy to know that the self-distance of a time-series will be zero, i.e., $D_{i, i}$=0. Besides, $D_{i, j}$ will be equal to $D_{j, i}$.

The $K$-means clustering algorithm \cite{kanungo2002efficient} will be applied to cluster the time-series using the calculated distance matrix $\mathbf D$. The original time-series $\mathbf X$ will be classified into $M$ groups, i.e., $\mathbf X_1$, $\mathbf X_2$, $\dots$, $\mathbf X_M$. After clustering the time-series dataset, unique degradation patterns are observed for each cluster.

\subsection{Construction of Features}\label{handfeature}
With classified degradation time-series $\mathbf X_m (m\in[1, M])$, features will be constructed correspondingly as the input of the following estimation models, rather than adopting the raw measurements. The min-max normalization is employed to overcome the negative influence of scale \cite{Heatpump}.

\subsubsection{Feature Engineering with Health Indicators} \label{healthindicator}
Multiple health indicators are constructed to capture distinct physical-meaning characteristics of the univariate signal \cite{JointRUL2021}, \cite{2022TASEXie}.

Six statistical time-domain features are adopted. Among them, $Mean$ smooths the signal by averaging samples with a sliding window, reflecting the trend of the signal. Standard deviation $SD$ indicates the deviation of a signal from its average, which is always used jointly with $Mean$. Root mean square error $RMSE$ will enlarge the values above average and shrink the values below the average. As a result, $RMSE$ is sensitive to severe degradation, while insensitive to incipient degradation. In light of this, $Mean$ and $SD$ are more suitable for early faults in comparison with $RMSE$. To examine the probability density function of the signal, kurtosis information measures the peak of the probability density function, which is a high-order statistics checking the impulse nature of a signal. The remaining features include shape factor ($SF$) and crest factor ($CF$), respectively.

\begin{figure}[!ht]
\centering
{\includegraphics[width=0.5\textwidth]{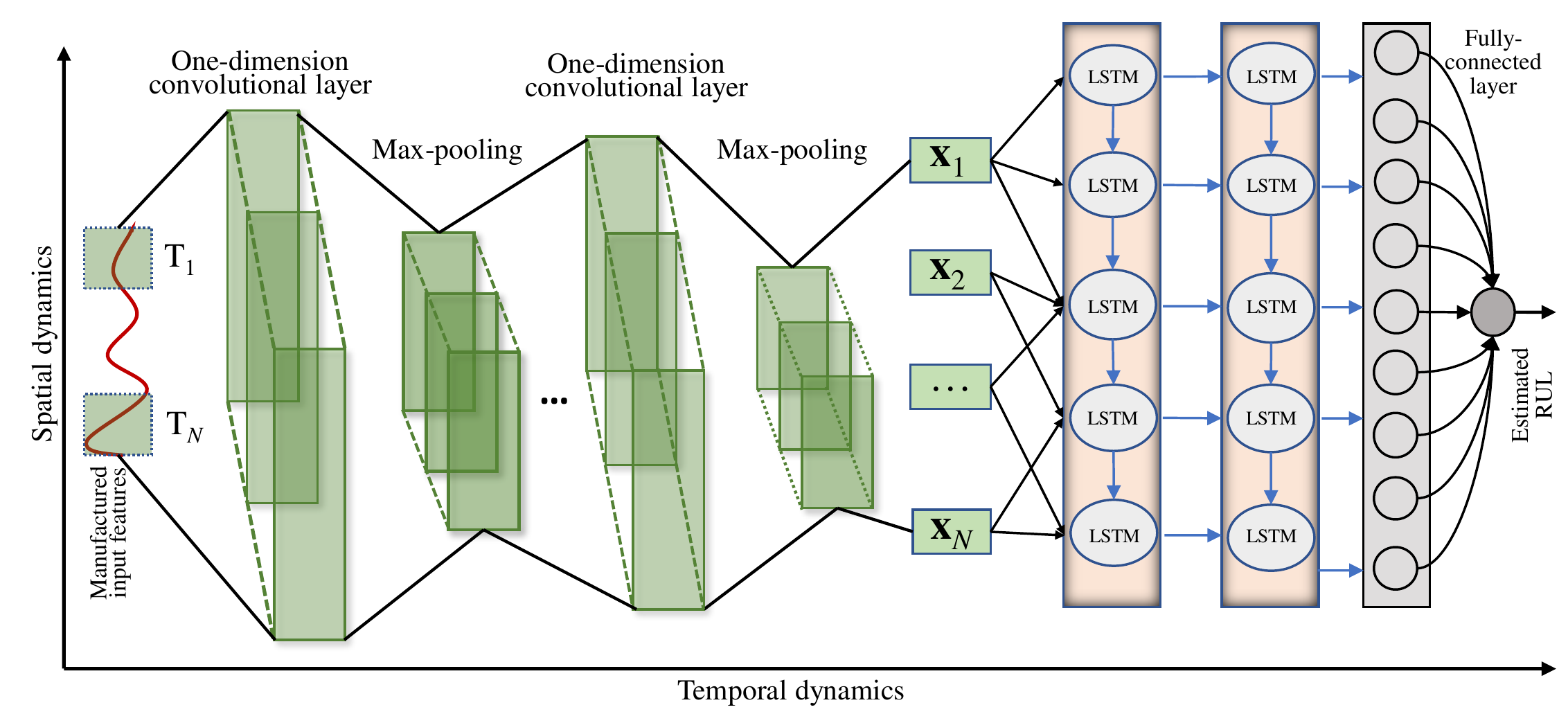}}
\caption{The illustration of the proposed RULNet network structure.}
\label{FIGTII6}
\end{figure}

With the above calculated six features, the original one-dimensional time-series is extended to seven dimensions, including the original measurements. As such, the new multivariate time-series of the $m^{th}$ cluster is denoted as $\mathbf X^{m}_{HI}$, where the symbol $HI$ indicates the health indicator.

\subsubsection{Feature Engineering with Self-Organizing Map}
As an unsupervised learning neural network, SOM aims to organize its neurons, i.e., map units, according to the nature of the input data. The map units are connected to adjacent neurons by a neighborhood relation, constructing a rectangular or hexagonal topology. A $n$-dimensional weight vector of the $i^{th}$ unit ${\mathbf u}_i = [u_{i, 1}, u_{i, 2}, \dots , u_{i, n}]$ connecting to the input data vector is employed to present each map unit.

The detailed training steps of SOM are summarized in \textbf {Algorithm 1}. Through iterative training, the weight vectors will be grouped into clusters depending on the calculated unified distance matrix (U-matrix). Usually, U-matrix is defined to visualize the distances between neighborhood units. For conciseness, the graphic presentation of SOM is not shown here, but the interested readers may refer to the literature dedicated to this topic \cite{SOM}.

Despite designed originally for sample clustering purposes, SOM has been applied for degradation feature extraction. The  idea behind this is that SOM will be trained first with time-series from normal operation to capture the data pattern before degradation patterns occur. Afterward, the quantitative degradation assessment is defined by calculation of minimum quantization error (MQE), which is the difference between the new measurement $\mathbf x_{new}$ and the corresponding best matching unit (BMU) of the trained SOM below,
\begin{equation}\label{eq:MQE}
MQE = ||\mathbf x_{new} - \mathbf u_{BMU}||
\end{equation}
where $\mathbf u_{BMU}$ stands for the weight vector of BMU.

For RUL estimation, the values of $MQE$ could be used as the inputs of preferred models, which are denoted as $\mathbf X^m_{SOM}$.

\subsubsection{Feature Engineering with Curve Fitting}
To better describe the evolution trend of the failed component, a curve fitting procedure is recommended to fit the time-domain features, including the indices $Mean$, $RMSE$, and $SD$. That is because the actual measurements of $Mean$, $RMSE$, and $SD$ features calculated from a run-to-failure time-series may suffer from severe fluctuations.

Actually, these fluctuations raise challenges for accurate RUL estimation. Curve fitting is an effective way to deal with the problem of fluctuations by combining failure mechanistic. Especially, Weibull failure rate function (WFRF) is capable of reflecting the fatigue strength and fatigue life of mechanical components, e.g., rolling bearing. Since only require to estimate four parameters, this function has been popular in mechanistic-based modeling. A modified version of WFRF is adopted here \cite{WeibullFunction}, which is given below,
\begin{equation}\label{eq:weibull}
\begin{array}{ll}
\lambda(k, \beta, \eta, a_1, c) = c + a_1 \frac{\beta}{\eta^\beta} k^{\beta-1}
\end{array}
\end{equation}
where $k$ is the cycle number of the time series, $\beta$ is the shape parameter, $\eta$ is the scale parameter, $a_1$ is a scale parameter, and $c$ is used to indicate the value when the time is 0.

Taking features of $Mean$, $RMSE$, and $SD$ as input, we can estimate the parameters $\beta$, $\eta$, $a_1$, and $c$ for corresponding functions using the least square regression.

WFRF fails to fit the decaying tail in degradation patterns observed in time domain features. Here, we introduce a decaying function with respect to the time, namely inverse failure rate function (IFRF), which is formulated as follows,
\begin{equation}\label{eq:inverse}
\begin{array}{ll}
F(k, \beta, a_2, a_3) = \frac{a_3}{(k+a_2)^\beta}
\end{array}
\end{equation}
where $k$ is the cycle number, $\beta$ is the shape parameter, $a_2$ is the shift parameter, and $a_3$ is the scale parameter.

\subsubsection{Comprehensive Comparisons between the Feature Engineering Ways}
Although the feature engineering ways mentioned above are popular, selecting a proper one for a specific task has not been available yet. On the basis of this, we comprehensively compare these ways from six aspects, as shown in Fig. \ref{FIGTII4}. Moreover, operational rules can be derived by evaluating the available domain knowledge and data. Each aspect is scored with three degrees, ranging from low and medium to high.

\begin{itemize}
\item \textit{Model complexity}: Health indicator is the easiest and fast way to extract features, while SOM costs the most computation resource since every sample will be computed with every node;
\item \textit{Domain knowledge}: Curve fitting requires an insightful understanding of the degradation process to fit the measurements to the prior distribution. In contrast, health indicator requires less domain knowledge compared with curve fitting. SOM requires the least domain knowledge by distinguishing normal samples from faulty samples;
\item \textit{Tuning parameters}: The sliding window length is the only tunable parameter of health indicators. The number of map units $N$, the learning rate $\alpha$, neighborhood function $H(\cdot)$, and the iteration stopping time have to be determined in SOM. According to Eqs. (3) and (4), more tunable parameters are needed to be optimized for curve fitting;

\item \textit{Training time}: Training time has a highly positive relationship with the model complexity. As a result, it consumes the most time to train SOM, while it spends the least time constructing the health indicators;
\item \textit{Causality analysis}: Causality evaluates the relevance between the constructed features and the real failure procedure. The higher relevance, the better RUL estimation result. Since curve fitting is developed with failure mechanistic, its causality is the highest. The health indicator may be the least relevant due to the inability to distinguish normal operation from failure pattern;
\item \textit{Robustness analysis}: Robustness evaluates the certainty of extracted features. Health indicators and curve-fitting own the highest robustness since they are only influenced by several tunable parameters. The random initialization of the SOM network weakens its robustness due to the common disadvantage of machine learning models.
\end{itemize}

$Remark$: To highlight the strengths of different features, the guidance rule for selecting proper feature engineering is illustrated in Fig. \ref{FIGTII5}. Concretely, if the failure pattern of the concerned component is known, curve fitting will be the first choice. Otherwise, we can further check data availability from normal operation conditions, which is essential to train SOM. Finally, health indicators have universal generality, which can be the fundamental way to extract features.

\subsection{The Network Structure of Proposed RULNet}
Network structure of the proposed RULNet as shown in Fig. \ref{FIGTII6} is systematically explained, which consists of input, convolutional operation, temporal operation, and the output.

Basically, the inputs of RULNet are the derived features mentioned in Section III.B, i.e., $\mathbf X^{m}_{HI}$ from the health indicators, $\mathbf X^{m}_{SOM}$ from SOM, or $\mathbf X^{m}_{CF}$ from the curve fitting, in the $m^{th}$ degradation pattern. Table II summarizes the detailed dimension information for each feature engineering, in which the dimension information of each feature has been specified. It is worth noticing that the same network configuration is adopted for all feature engineering ways, as the input layer with one-dimension convolutional operation is insensitive to the feature dimension. More details about each part are given as follows.

\subsubsection{Convolutional Feature Layer}\label{CNNLayer}
Serial convolutional units are stacked together through the convolution of features, and hence achieving high-level and deep representation of the features. Here, a single convolutional unit consists of a 1D-convolution layer and a 1D-max-pooling layer. Assuming that $\mathbf d^{(l-1)}$ and $\mathbf d^{l}$ stand for the input and the output feature maps of the $l$th 1D-convolution layer, respectively. Input to the $l^{th}$ layer is the output of the $(l-1)^{th}$ layer. Since there are multiple channels for an output feature map from a layer, we denote the $j^{th}$ feature channel of the layer $l$ as $\mathbf d_j^{l}$. Operation of the 1D-convolution layer can be formulated by,

\begin{equation}\label{eq:conv}
\mathbf d_j^{l}=R\left( \sum_{i} \mathbf d_i^{(l-1)}*\vec{\mathbf w}_{i,j}^{l} + \mathbf b_j^{l}\right)
\end{equation}
where $*$ denotes the convolution operator; $\vec{\mathbf w}_{i,j}^{l}$ and $\mathbf b_j^{l}$ represents the one dimensional weight of the convolution kernel and the bias of the $j^{th}$ feature map of the $l^{th}$ layer, respectively; and $R(\cdot)$ is a activation function with a rectified linear unit.

\begin{table}[!htb]
\centering
\caption{The dimensions of each constructed features.}
\renewcommand{\arraystretch}{1}
\scriptsize
\begin{tabular}{p{3cm} p{4.5cm}}
	\midrule
    \midrule
	\textbf{Feature name} & \textbf{Dimension and descriptions} \\
	\midrule
    Feature engineering with health indicators & A vector with seven dimensions, including six indicators and the one-dimensional original measurement\\
    \midrule
    Features engineering with self-organizing map & One dimensional vector with minimum quantization error\\
    \hline
    Features engineering with curve fitting & A vector with three dimensions manually fitted by Eqs. (4) or (5) using mean, root mean square error, and stand variation \\
	\midrule
    \midrule
\end{tabular}
\end{table}

Pooling layers follow with the convolution layers as a progressive mechanism to reduce the spatial size of the feature representations in CNN layers. This improves computational efficiency and enables the deep structure by reducing the number of parameters with sharing coefficients. Max-pooling is the most recommended way of sub-sampling in CNNs, which is given below,
\begin{equation}\label{eq:max}
\mathbf d_{j, i}^{l}= \max \left(\mathbf d_{j_{i, b}}^{l}\right),
\end{equation}
where $\mathbf d_{j, i}^{l}$ represents the $i^{th}$ element of feature map $\mathbf d_j^{l}$ and $\mathbf d_{j_{i, b}}^{l}$ is the set of values in the 1D-neighborhood of $\mathbf d_{j, i}^{l}$.

A fully connected layer will follow a series of convolutional and max-pooling layers to enable the function of LSTM layers. Let $\mathbf F_{j}$ be the $j^{th}$ flattened feature map of the last max-pooling layer, where $j \in [1, 2, \dots , Q]$. The set of flattened feature maps in the last layer can be represented as $\mathbf F=[\mathbf F_1, \mathbf F_2, \dots, \mathbf F_Q]$. This flattened layer passes through a fully-connected neural network as follows,
\begin{equation}
\mathbf x = F\left( \mathbf W_{c} \mathbf F+ \mathbf b_{c}\right)
\end{equation}
where $\mathbf x$ represents the output of fully-connected layer; $\mathbf W_{c}$ and $\mathbf b_{c}$ represent the transformation weights and bias of the fully-connected layer, respectively; and softmax function is usually selected for $F(\cdot)$.

\subsubsection{Temporal Correlation Layer}\label{lstm}
The learning ability regarding temporal correlation is ensured via LSTM, which is a sequence to sequence learning mechanism to relate an entire sequence with target state outputs in a state-space manner.

The basic architecture of LSTM consists of dozens of LSTM cells with memory capability. The LSTM cell is designed to overcome the vanishing gradient in the nodes of traditional RNN by introducing three gate functions and a memory state. The information flow of the LSTM cell is regulated as follows. For an easy understanding of the temporal correlation, the time index $k$ is added here. First, hidden information $\mathbf h_{k-1}$ at time $k-1$ and input information $\mathbf x_k$ calculated from Eq. (7) at time $k$ are simultaneously imported into input gate $i$, forget gate $f$, and output gate $o$. The corresponding gating variables are obtained as follows,
\begin{equation}
\begin{aligned}
\mathbf{i}_{k} &=\delta(\mathbf{W}_{i} \mathbf{h}_{k-1}+\mathbf{U}_{i} \mathbf{x}_{k}+\mathbf{b}_{i}) \\
\mathbf{f}_{k} &=\delta(\mathbf{W}_{f} \mathbf{h}_{k-1}+\mathbf{U}_{f} \mathbf{x}_{k}+\mathbf{b}_{f}) \\
\mathbf{o}_{k} &=\delta(\mathbf{W}_{o} \mathbf{h}_{k-1}+\mathbf{U}_{o} \mathbf{x}_{k}+\mathbf{b}_{o})
\end{aligned}
\end{equation}
where $\mathbf i_k$, $\mathbf f_k$, and $\mathbf o_k$ are outputs of input gate, forget gate, and output gate, respectively; $\mathbf W_i$, $\mathbf U_i$, $\mathbf W_f$, $\mathbf U_f$, $\mathbf W_o$, and $\mathbf U_o$ are weighting matrixes of the above three gates, respectively; $\mathbf b_i$, $\mathbf b_f$, and $\mathbf b_o$ are biases of three gates, respectively; $\delta(\cdot)$ is element-wise sigmoid function.

Next, a candidate cell memory state is calculated below,
\begin{equation}
\tilde{\mathbf{c}}_{k}=\tanh(\mathbf{W}_{c} \mathbf{h}_{k-1}+\mathbf{U}_{c} \mathbf{x}_{k}+\mathbf{b}_{c})
\end{equation}
where $\mathbf W_c$ and $\mathbf U_c$ are weighting matrixes, $\mathbf b_c$ is the bias, and $\tanh(\cdot)$ is element-wise hyperbolic tangent function.

Then final cell state $\mathbf c_k$ is updated with forgetting information $\mathbf f_k$ and input information $\mathbf i_k$ below,
\begin{equation}
\mathbf{c}_{k}=\mathbf{f}_{k} \circ \mathbf{c}_{k-1}+\mathbf{i}_{k} \circ \tilde{\mathbf{c}}_{k}
\end{equation}
where operator $\circ$ denotes the element-wise product function of two vectors.

Finally, hidden information $\mathbf h_k$ at time $k$ is updated with $\mathbf o_k$ and $\mathbf c_k$ as follows,
\begin{equation}
\mathbf{h}_{k}=\mathbf{o}_{k} \circ \tanh (\mathbf{c}_{k})
\end{equation}

By repeating the above procedures with time evolution, temporal prediction ability is owned by the LSTM network.

The final output of RUL estimation is achieved by following LSTM layers with a fully-connected layer. Let $\mathbf O$ be the flattened sequence of the final LSTM layer output $\mathbf o_k$. After randomly dropping out partial neurons in the fully-connected layer to avoid over-fitting, the remaining neurons are connected to a single node to output the RUL estimation,
\begin{equation}
RUL = F\left( \mathbf W_{L} \mathbf O+ \mathbf b_{L}\right)
\end{equation}
where $\mathbf W_{L}$ and $\mathbf b_{L}$ represent the weights and bias of the fully-connected layer after dropping out, respectively.

\subsection{Online Estimation with the Hybrid RULNet Model}
With the classified patterns and the developed RULNet model, online estimation could be conducted when the online run-to-failure time-series are available. The basic steps for online estimation are summarized as follows:

(1) Obtain an online run-to-failure data denoted as $\mathbf x_{online}$.

(2) Calculate the DTW distance between $\mathbf x_{online}$ and the center of each degradation pattern, and judge the belonging of pattern clusters for the run-to-failure time-series $\mathbf x_{online}$.

(3) Adopt a specific feature according to the guidance selected from the offline modeling.

(4) Feed the constructed feature into the well-developed RULNet model;

(5) Obtain the online RUL estimation results with the hybrid RULNet model.

\section{Experimental Results and Discussion}
In this section, the efficacy of the proposed method has been verified through comprehensive comparisons with existing works through the practical dataset described in Section II.

%

\subsection{Clustering for Run-to-Failure Time-Series}
The necessity of grouping lies in that we need to identify distinct patterns in run-to-failure time-series to apply unique RUL estimation models for each pattern separately. Moreover, mixing the run-to-failure time-series with different degradation patterns might result in false RUL estimation. Furthermore, we can have a better insight into how run-to-failure data behaves according to the clusters.

\begin{figure*}[!ht]
	\centering
	\subfigure[]	
	{
	\begin{minipage}[t]{0.3\linewidth}
	\centering
	\includegraphics[width=5cm]{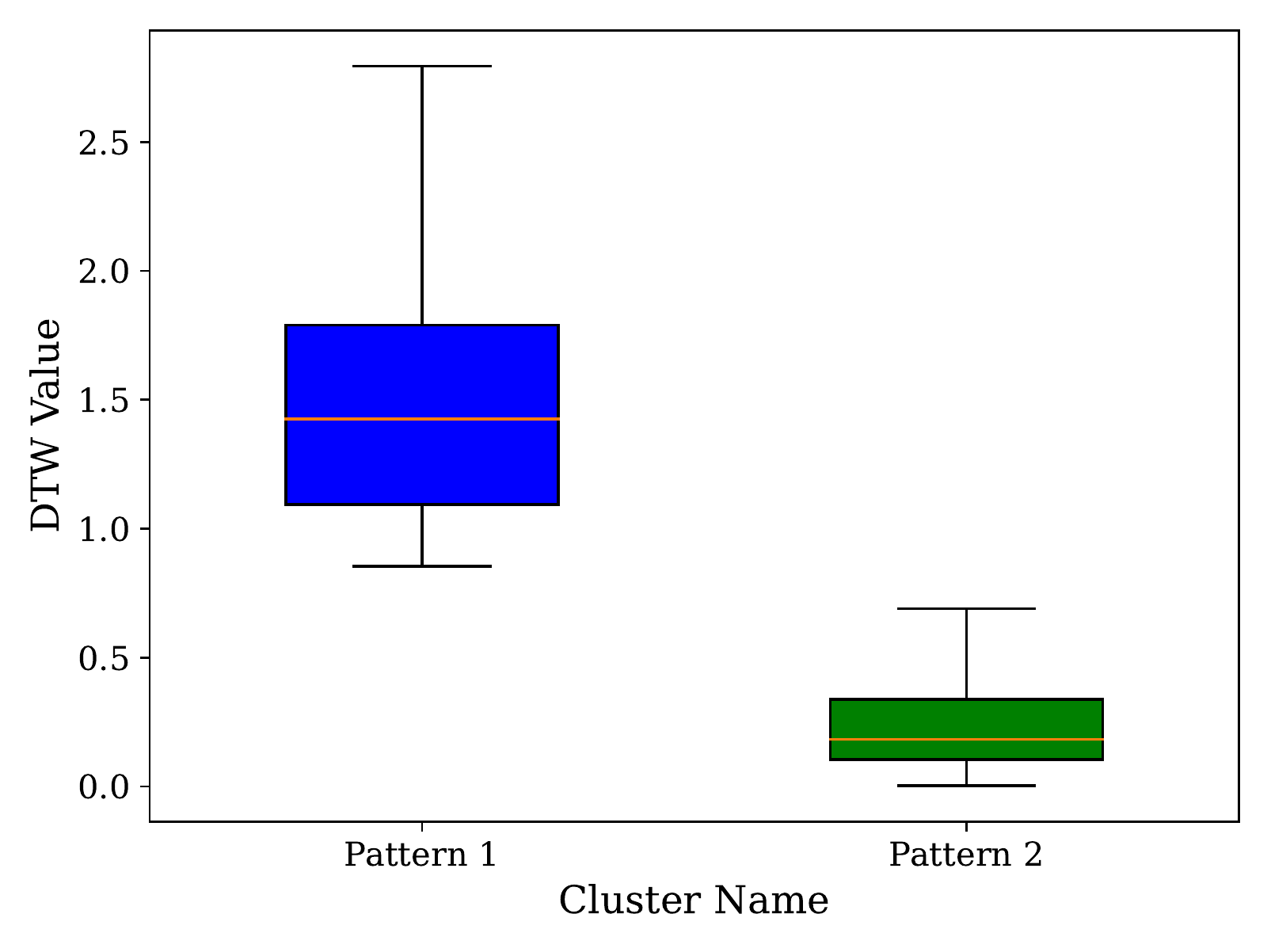}
	\end{minipage}
	}
	\subfigure[]
	{
	\begin{minipage}[t]{0.3\linewidth}
	\centering
	\includegraphics[width=5cm]{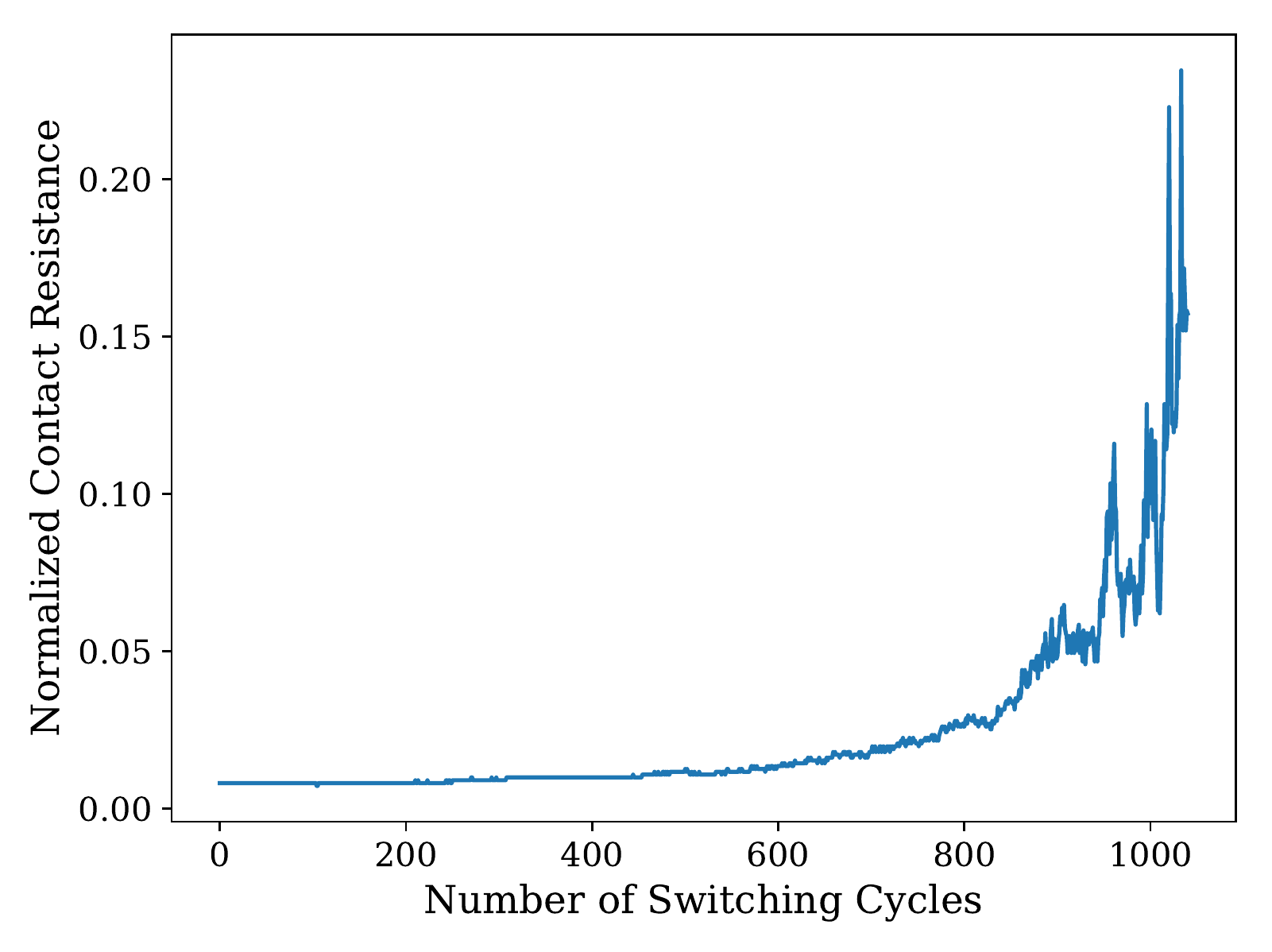}
	\end{minipage}
	}
	\subfigure[]	
	{
	\begin{minipage}[t]{0.3\linewidth}
	\centering
	\includegraphics[width=5cm]{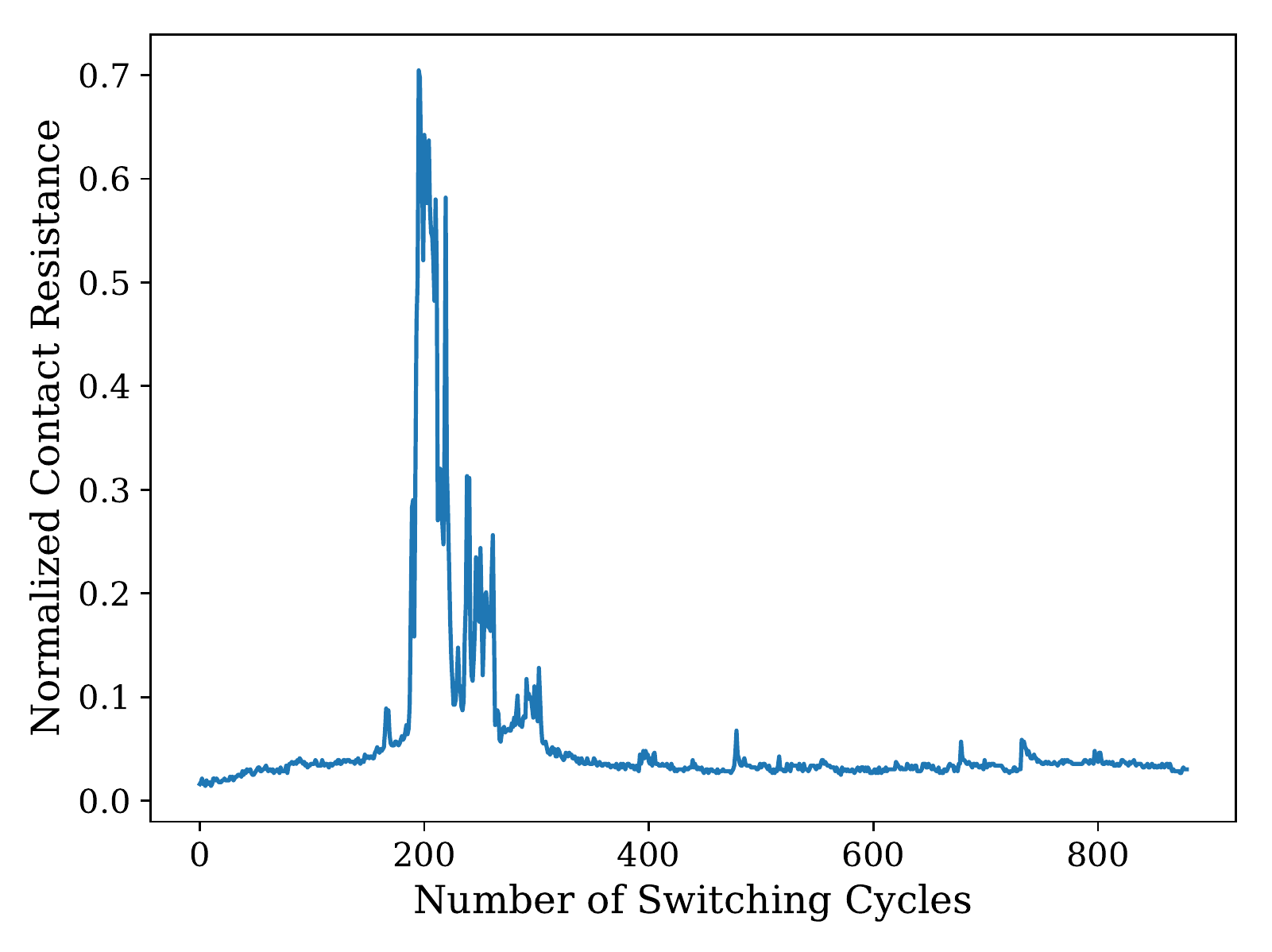}
	\end{minipage}
	}
	\caption{The illustration of (a) Clustering of run-to-failure time-series from training data with similarity measured by DTW for each pattern, (b) one typical failure sample in Pattern 1, and (c) one typical failure sample in Pattern 2.}
	\label{FIGTII7}
\end{figure*}

\begin{table*}[!htb]
	\caption{Network configuration of RULNet}
	\renewcommand{\arraystretch}{1}
	\label{tb:system_model}
    \scriptsize
    \centering
	\begin{tabular}{p{1.2cm} p{1.3cm} p{3cm} p{1.5cm} }
		\midrule
        \midrule
		\textbf{Layer No.} & \textbf{Name} & \textbf{Parameter Configuration} & \textbf{Output Shape}\\
		\midrule
		{Layer 1} & 1D-CNN & Filters=18, Kernel\_Size=2, Strides=1, Padding=same & $100 \times 18$ \\
        \midrule
		{Layer 2} & Max-pooling & Pool\_Size=2, Strides=2, Padding=same  & $50 \times 18$ \\
		\midrule
		{Layer 3} & 1D-CNN & Filters=36, Kernel\_Size=2, Strides=1, Padding=same & $50 \times 36$ \\
		\midrule
		{Layer 4} & Max-pooling & Pool\_Size=2, Strides=2, Padding=same &  $25 \times 36$ \\
		\midrule
		{Layer 5} & 1D-CNN & Filters=72, Kernel\_Size=2, Strides=1, Padding=same & $25 \times 72$ \\
		\midrule
		{Layer 6} & Max-pooling & Pool\_Size=2, Strides=2, Padding=same & $13 \times 72$ \\
		\midrule
		{Layer 7} & FNN & Units=sequence\_length*channels  &  $700$ \\
		\midrule
		{Layer 8} & Dropout & Dropout\_Probability=0.2  &  $700$ \\
		\midrule
		{Layer 9} & LSTM & Layer\_Size=channels*3  &  $100 \times 21 $ \\
		\midrule
		{Layer10} & Dropout & Dropout\_Probability=0.2  &  $100 \times 21$ \\
		\midrule
		{Layer 11} & LSTM & Layer\_Size=channels*3    &    $21$ \\
		\midrule
		{Layer 12} & Dropout & Dropout\_Probability=0.2   &   $21$ \\
		\midrule
		{Layer 13} & FNN & Layer\_Size=50 & 10 \\
		\midrule
		{Layer 14} & Dropout & Dropout \_Probability=0.2   &  10 \\
		\midrule
		{Layer 15} & FNN & Layer\_Size=1 (output layer)  &  1 \\
		\midrule
        \midrule
	\end{tabular}
\end{table*}

\begin{table}[ht]
\centering
\scriptsize
\label{tb:parameters_patterns1}
\renewcommand{\arraystretch}{1.5}
\caption{Estimated WFRF parameters for run-to-failure time-series in Pattern 1}
\begin{tabular}{p{0.9cm} p{2cm} p{1.3cm} p{1.3cm} p{1.3cm}}
	\hline
    \hline
	\multirow{2}{*}{\textbf {Index}} & \multicolumn{4}{c}{\textbf{Parameter}} \\
	\cline{2-5}
	\textbf {} & \textbf {$c$} & \textbf {$\eta$}  & \textbf {$\beta$} & \textbf {$k$}\\
	\hline
	$Mean$ & $1.027 \times 10^{-2}$   & $1.060 \times 10^{0}$  & $9.545 \times 10^{0}$ & $1.290 \times 10^{1}$\\
        \hline
	$RMSE$ & $1.034 \times 10^{-2}$   & $1.069 \times 10^{0}$  & $9.715 \times 10^{0}$ & $1.285 \times 10^{1}$\\
	\hline
	$SD$   & $3.065 \times 10^{-4}$   & $1.272 \times 10^{0}$  & $1.839 \times 10^{1}$ & $1.313 \times 10^{1}$\\
	\hline
    \hline
\end{tabular}
\end{table}

\begin{figure}[!ht]
	\subfigure[]
	{
	\begin{minipage}[t]{1\linewidth}
	\centering
	\includegraphics[width=6.5cm]{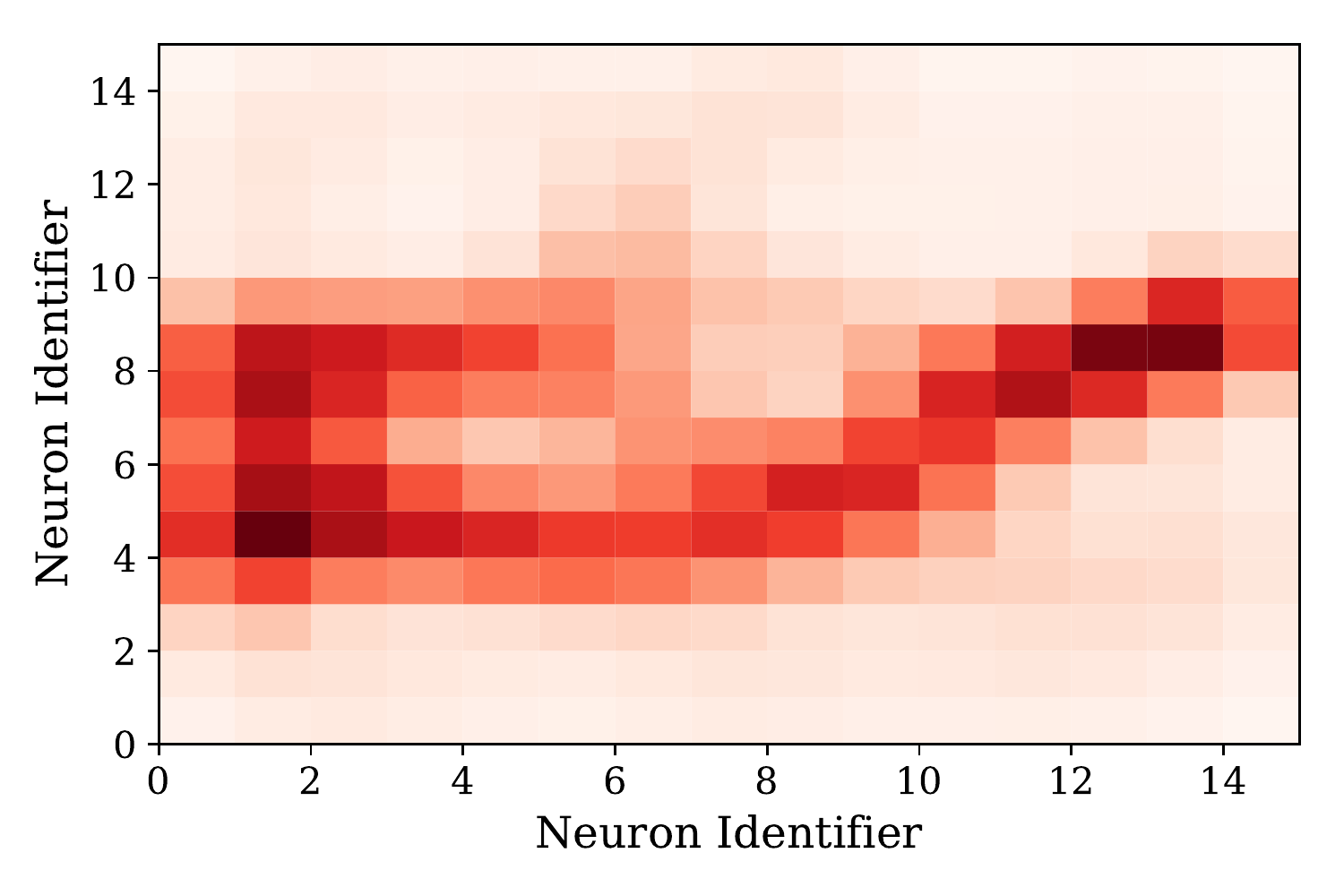}
	\end{minipage}
	}

	\subfigure[]	
	{
	\begin{minipage}[t]{1\linewidth}
	\centering
	\includegraphics[width=6.5cm]{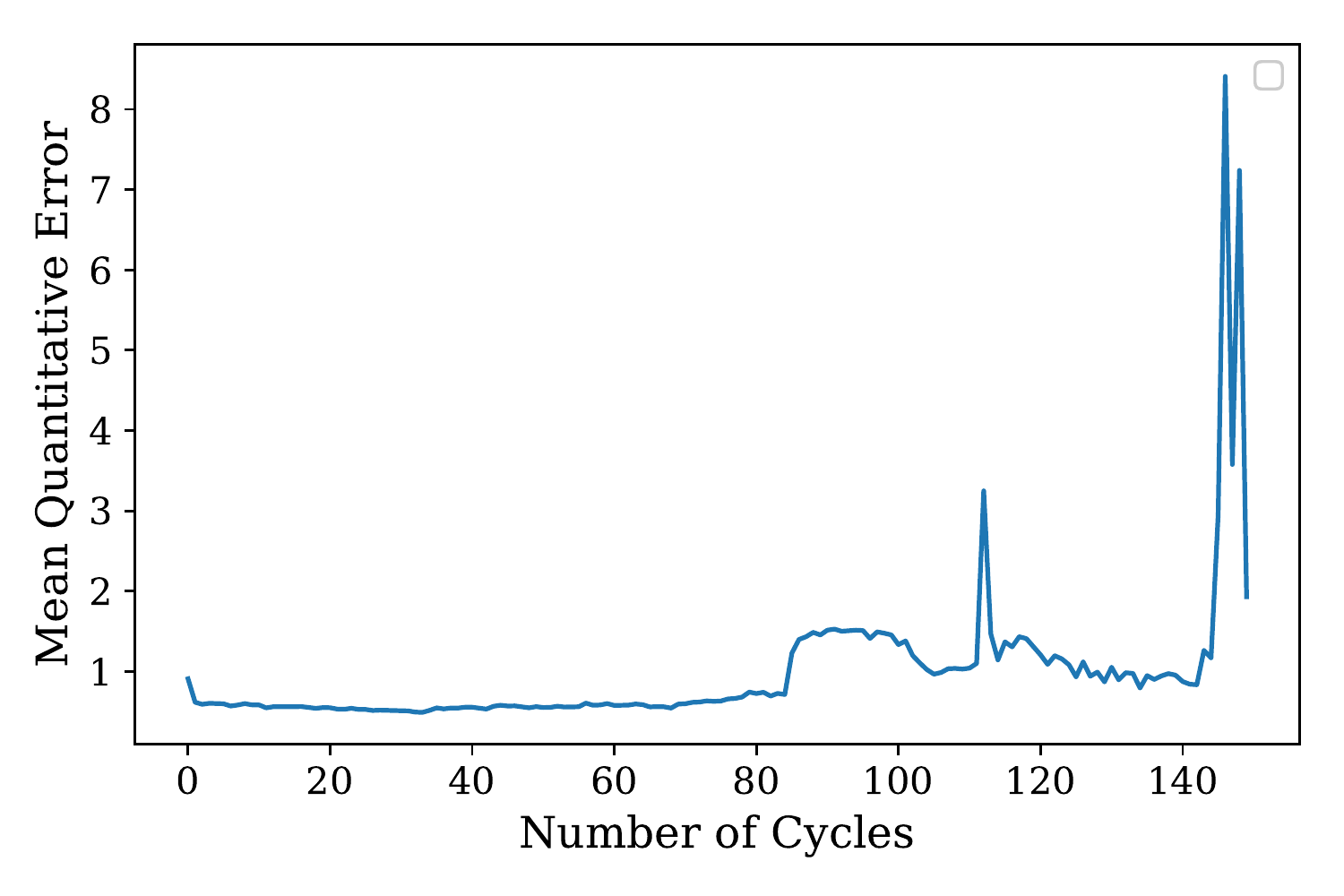}
	\end{minipage}
	}	
	\caption{The trained SOM in Pattern 1 with (a) U-matrix and (b) Calculated mean quantitative error.}
	\label{FIGTII8}
\end{figure}

\begin{table}[!ht]
\centering
\caption{Performance comparison between the proposed RULNet and its counterparts concerning the index $RMSE$.}
\scriptsize
\renewcommand{\arraystretch}{1.5}
\begin{tabular}{c|c|c|c}
	\hline
    \hline
	{\textbf{Feature Engineering}} & {\textbf{Network}} & {\textbf{Pattern 1}} & {\textbf{Pattern 2}} \\
	\hline
    \multirow{2}{*}{{Raw data}} & LSTM  & 27.28 & 28.61 \\
	\cline{2-4}
	& CNN & 27.8 & 26.76 \\
    \hline
	\multirow{3}{*}{{Features with health indicators}} & LSTM  & 23.68 & 23.95 \\
	\cline{2-4}
	& CNN &  22.11 & 19.40  \\
	\cline{2-4}
	& RULNet (Proposed)  & \textbf{22.09} & \textbf{18.50} \\
	\hline
	\multirow{3}{*}{{Features with SOM}} & LSTM & 22.46 & NA \\
	\cline{2-4}
	& CNN & 21.90  & NA  \\
	\cline{2-4}
	& RULNet (Proposed) & \textbf{21.86} & NA  \\
	\hline
	\multirow{3}{*}{{Features with curve fitting}} & LSTM & 21.53  & {21.19} \\
	\cline{2-4}
	& CNN & \textbf{20.47} & 20.29  \\
	\cline{2-4}
	& RULNet (Proposed) & {20.61} & \textbf{20.08} \\
	\hline
    \hline
\end{tabular}
\begin{tablenotes}
	\item[1] NA means not applicable, as SOM feature is not applicable to Pattern 2.
\end{tablenotes}
\label{tb:perf_table2}
\end{table}

70$\%$ data collected from Section II are used as training data for clustering, and then the size of training data is 174. Time-series in failure reed relays are classified into two clusters according to the given procedure in Section III.A, as shown in Fig. \ref{FIGTII7} with Application Programming Interfaces from Tslearn python package. As shown in Fig. \ref{FIGTII7}(a), all time-series are grouped into two clusters by measuring their distances of DTW with $K$-means. Specifically, the number of time-series classified into the first cluster is 151, and the second cluster consists of the number of 23 time-series. The first cluster in Fig. \ref{FIGTII7}(b) shows an increasing pattern of the contact resistance, referred to as Pattern 1. The second cluster in Fig. \ref{FIGTII7}(c) shows a signal that increases to a maximum in its first half and then decreases gradually, named Pattern 2.

\begin{figure*}[!ht]
	\centering
	\subfigure[]
	{
	\begin{minipage}[t]{0.3\linewidth}
	\centering
	\includegraphics[width=5.5cm]{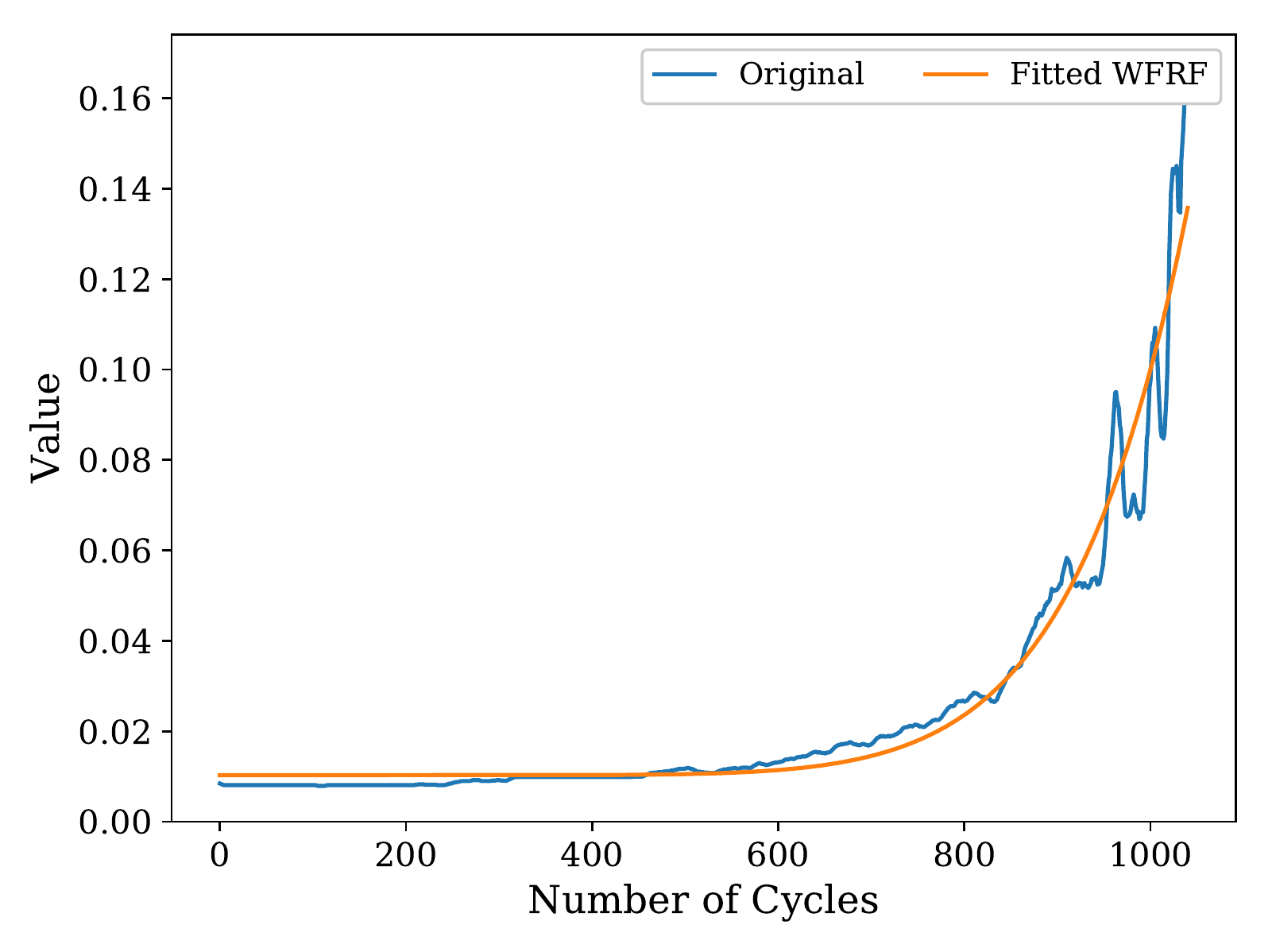}
	\end{minipage}
	}
	\subfigure[]
	{
	\begin{minipage}[t]{0.3\linewidth}
	\centering
	\includegraphics[width=5.5cm]{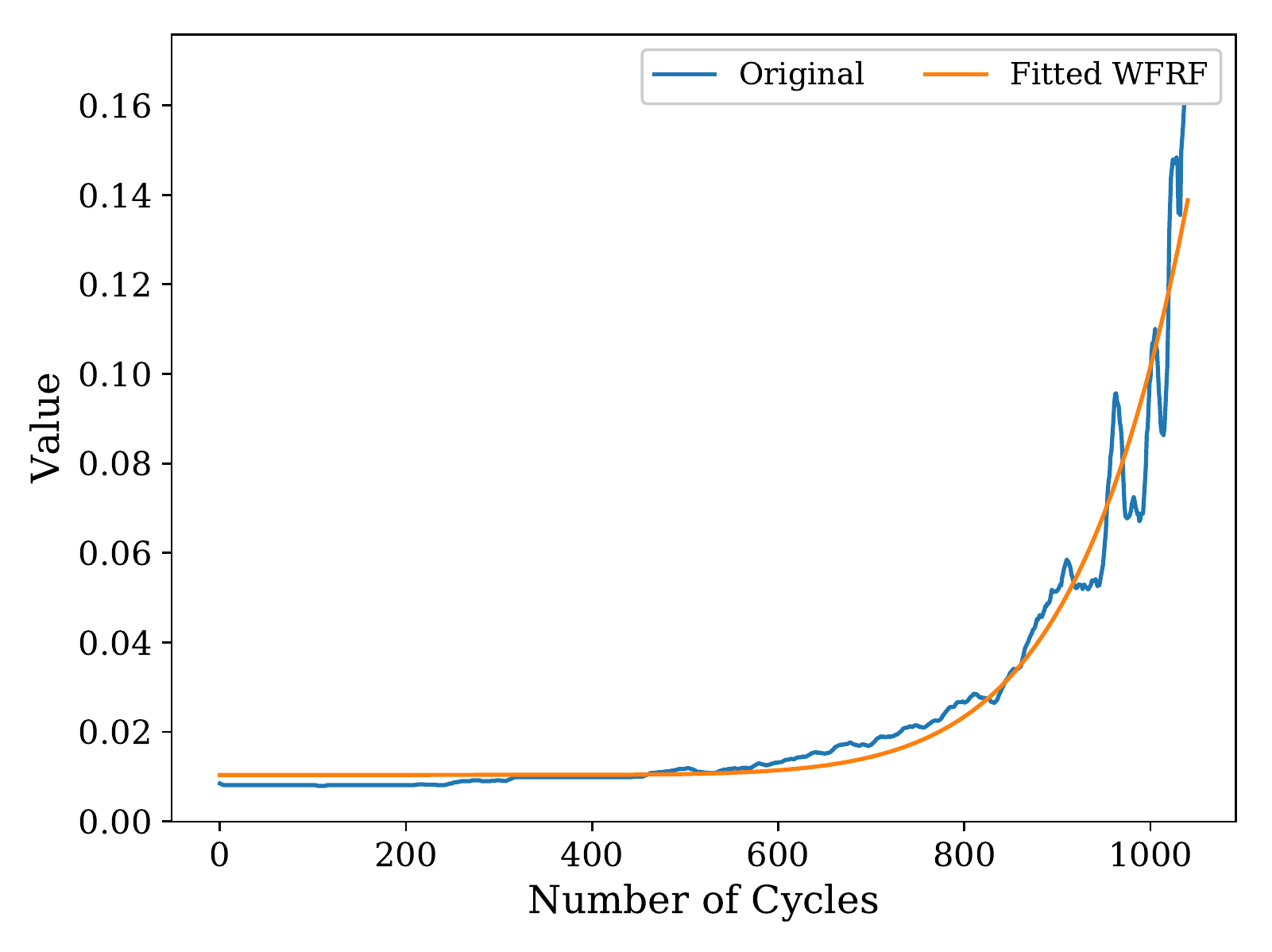}
	\end{minipage}
	}
	\subfigure[]	
	{
	\begin{minipage}[t]{0.3\linewidth}
	\centering
	\includegraphics[width=5.5cm]{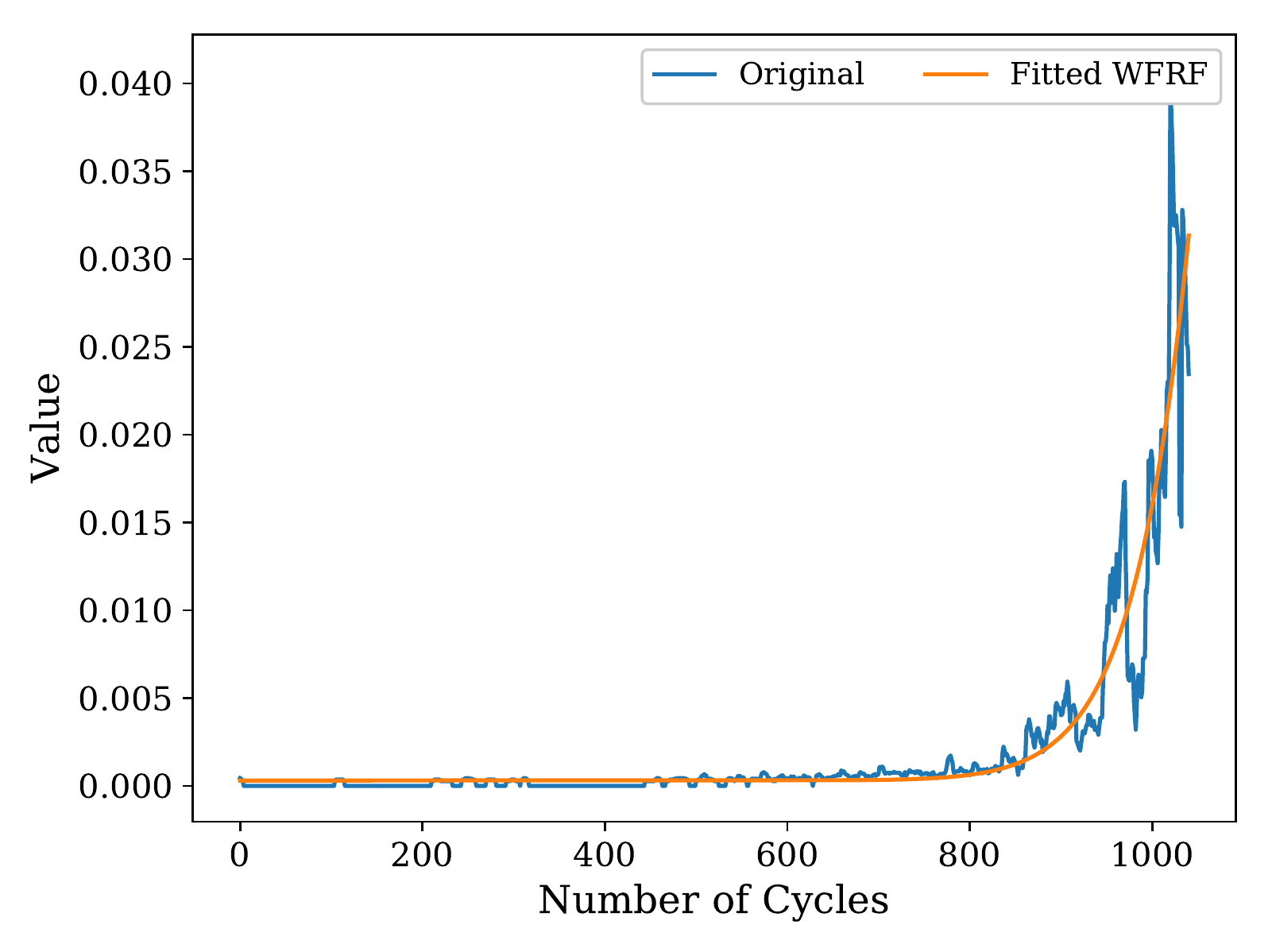}
	\end{minipage}
	}	
	\caption{Curve fitting for a run-to-failure time-series in Pattern 1 using (a) Mean, (b) standard deviation, and (c) RMSE.}
	\label{FIGTII9}
\end{figure*}

\begin{figure*}[!ht]
	\centering
	\subfigure[]
	{
	\begin{minipage}[t]{0.3\linewidth}
	\centering
	\includegraphics[width=5.5cm]{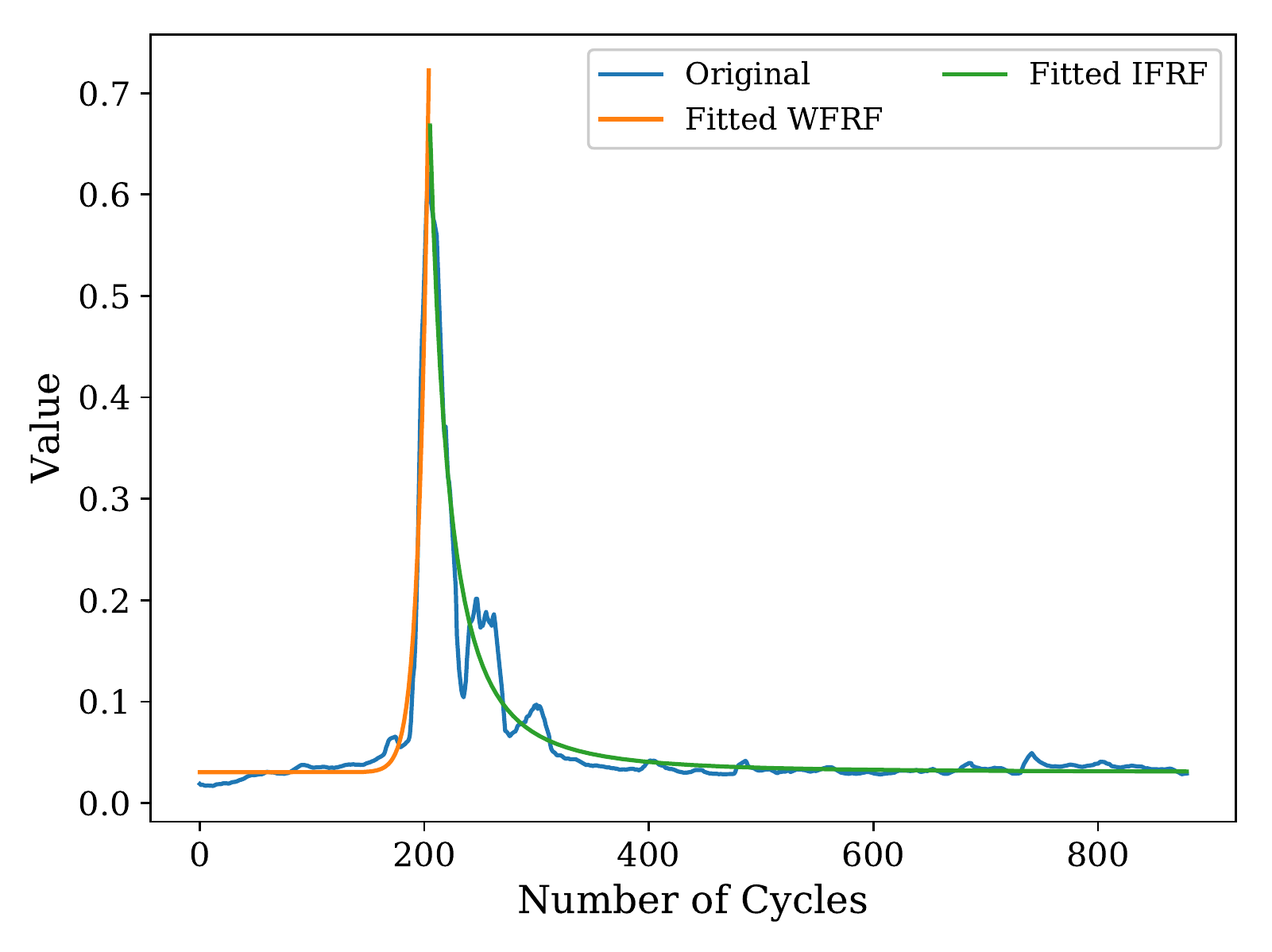}
	\end{minipage}
	}
	\subfigure[]
	{
	\begin{minipage}[t]{0.3\linewidth}
	\centering
	\includegraphics[width=5.5cm]{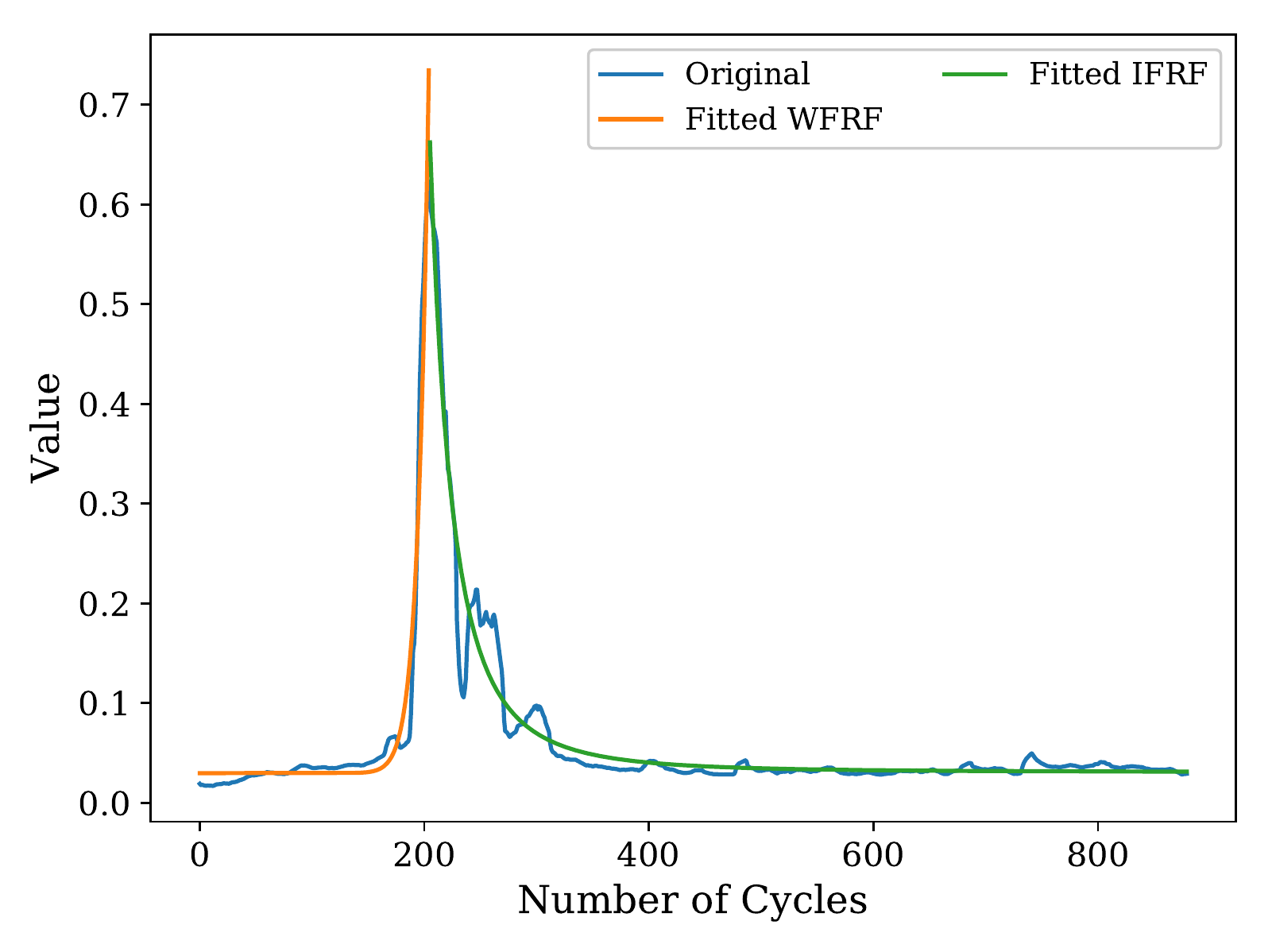}
	\end{minipage}
	}
	\subfigure[]	
	{
	\begin{minipage}[t]{0.3\linewidth}
	\centering
	\includegraphics[width=5.5cm]{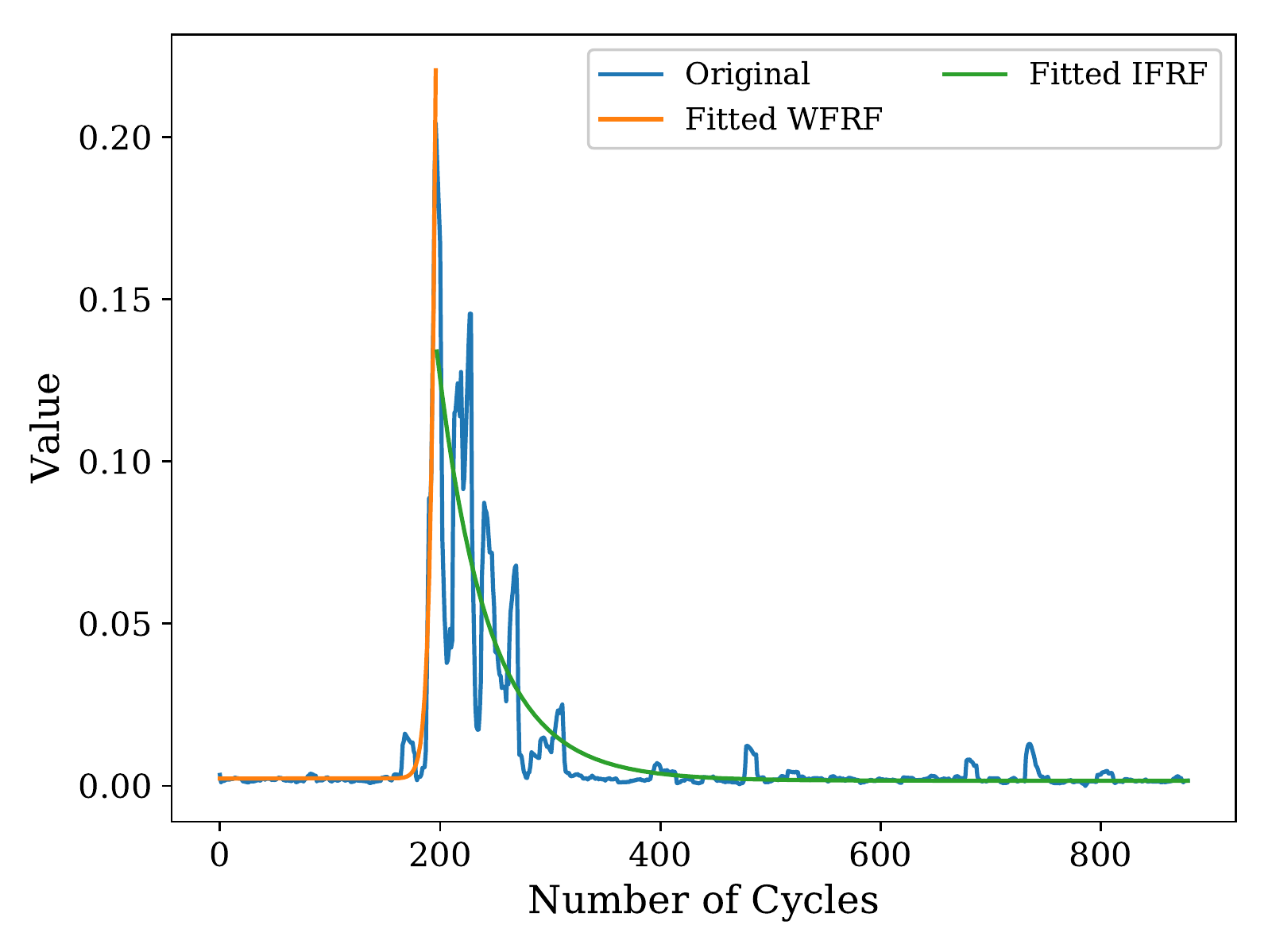}
	\end{minipage}
	}	
	\caption{Curve fitting for a run-to-failure time-series in Pattern 2 using (a) Mean, (b) standard deviation, and (c) RMSE.}
	\label{FIGTII10}
\end{figure*}

\subsection{Construction of Features}
According to the clustering results of degradation patterns given in the last subsection, the run-to-failure time-series have two degradation patterns, each of which is treated separately in the following feature ways.

\subsubsection{Feature Engineering with Health Indicators}
Six features are extracted as mentioned in Section III.B.1 to construct the multivariate time-series to train the RULNet model. This feature construction is conducted per run-to-failure time-series and per degradation pattern.

\subsubsection{Feature Engineering with SOM} Normal time-series are collected for training the SOM neural network. For easy understanding, the U-Matrix calculated with Algorithm 1 has been demonstrated in Fig. \ref{FIGTII8}(a). The more the color difference between neurons in U-matrix, the higher the difference of associated weights is. Normally, SOM works as an anomaly detector by outputting lower MQE values for normal operating condition data and higher MQE values for faulty operating condition data. Fig. \ref{FIGTII8}(b) shows the generated MQE values, demonstrating higher MQE values at later cycles. The calculated MQE values are used to train the RULNet model and the other two baseline models. This is the idea behind training SOM using normal time-series data and generating MQE series for run-to-failure data from Pattern 1. Since lacking of normal time-series in Pattern 2, MQE values cannot be generated for Pattern 2.

\subsubsection{Feature Engineering with Curve Fitting} With the normalized measurements of contact resistance from run-to-failure time-series, three time-domain features are calculated, namely $Mean$, $SD$, and $RMSE$. Then predefined functions, i.e., Eqs. (\ref{eq:weibull}) and (\ref{eq:inverse}), are fitted for these constructed features and normalized resistance series using the least square regression algorithm. The data in Pattern 1 are used to estimate the parameters of WFRF mentioned in Eq. (\ref{eq:weibull}). For the data in Pattern 2, the samplings in the first half until the maximum contact resistance is fitted into the WFRF, and the second half is fitted to the IFRF mentioned in Eq. (\ref{eq:inverse}). Example curve fitting scenario for a run-to-failure in Pattern 1 is illustrated in Fig. \ref{FIGTII9}. Fig. \ref{FIGTII10} shows curves fitted for a run-to-failure time-series in Pattern 2. Parameters of WFRF and IFRF corresponding to these two time-series are calculated using the least square algorithm are specified in Tables II and III.

\subsection{The Network Configuration of Proposed RULNet}
The specific architecture of the proposed approach consists of 15 layers, among which the first three layers are convolutional neural network. The first CNN layer consists of 18 filters, the second CNN consists of 36 filters, and the final CNN layer consists of 72 filters. As shown in Table \ref{tb:system_model}, every CNN layer is followed by a 1D-max-pooling layer, and the size of two kernels are used in every convolution and pooling operation. A fully-connected neural network (FNN) layer with dropout regularization has been introduced to connect the CNN layer to the LSTM layers. Since the sequence length is equal to 100, the input to the first CNN layer is represented as $\{ \mathbf X_n^k|k=k_i,...,k_i+99\}$, where $k_i$ is the end time step of the previous input of the first CNN layer. The LSTM layer size has been determined empirically. The final FNN layer works as a regression layer to estimate the RUL values.

\begin{table*}[!ht]
\caption{Estimated WFRF and IFRF parameters for run-to-failure time-series in Pattern 2}
\scriptsize
\centering
\label{tb:parameters_patterns2}
\renewcommand{\arraystretch}{1.5}
\begin{tabular}{c c c c c c c c}
	\hline
    \hline
	\multirow{2}{*}{\textbf {Parameter}}  & \multicolumn{4}{c}{\textbf {WFRF}} & \multicolumn{3}{c}{\textbf {IFRF}} \\
	\cline{2-8}
	\textbf {} & \textbf {$c$} & \textbf {$\eta$}  & \textbf {$\beta$} & \textbf {$k$} & \textbf {$\beta$} & \textbf {$a$} & \textbf {$k$}\\
	\hline
	$Mean$ & $3.034\times10^{-2}$ & $3.715\times10^{1}$  & $2.393\times10^{1}$  &  $2.436\times10^{1}$ & $7.203\times10^{-1}$  & $1.008\times10^{0}$  & $9.089\times10^{-1}$ \\
	\hline
	$RMS$ & $2.982\times10^{-2}$  & $6.086\times10^{1}$  & $2.216\times10^{1}$  &  $2.515\times10^{1}$ & $8.298\times10^{-1}$  & $1.172\times10^{0}$  & $9.486\times10^{-1}$ \\
	\hline
	$SD$ & $2.268\times10^{-3}$ & $6.216\times10^{0}$  & $4.188\times10^{1}$  &  $2.172\times10^{1}$ & $2.501\times10^{0}$  & $4.781\times10^{0}$  & $1.799\times10^{0}$ \\
	\hline
    \hline
\end{tabular}
\end{table*}

\begin{figure*}[!ht]
	\centering
	\subfigure[]
	{
	\begin{minipage}[t]{0.3\linewidth}
	\centering
	\includegraphics[width=5.5cm]{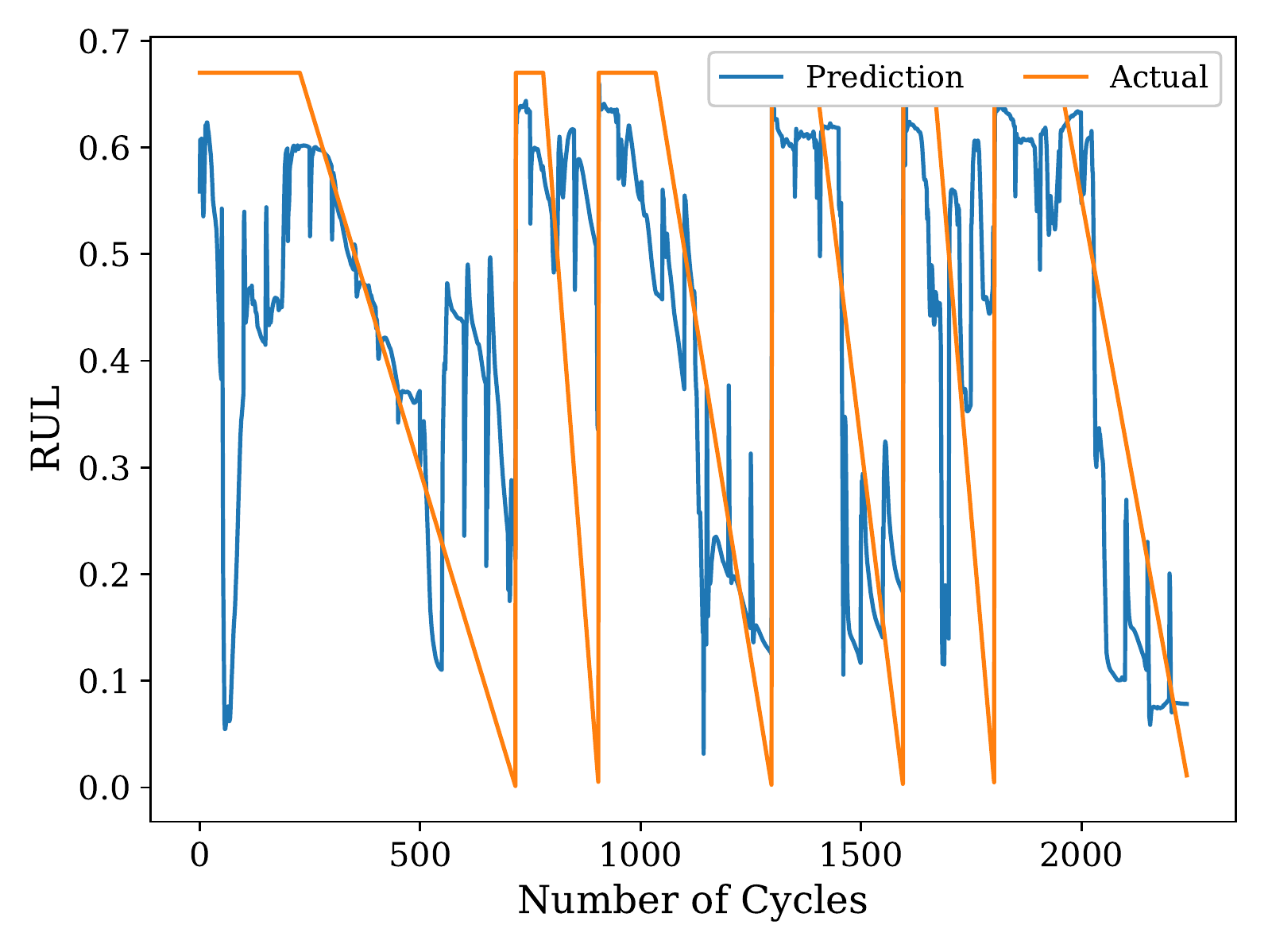}
	\end{minipage}
	}
	\subfigure[]
	{
	\begin{minipage}[t]{0.3\linewidth}
	\centering
	\includegraphics[width=5.5cm]{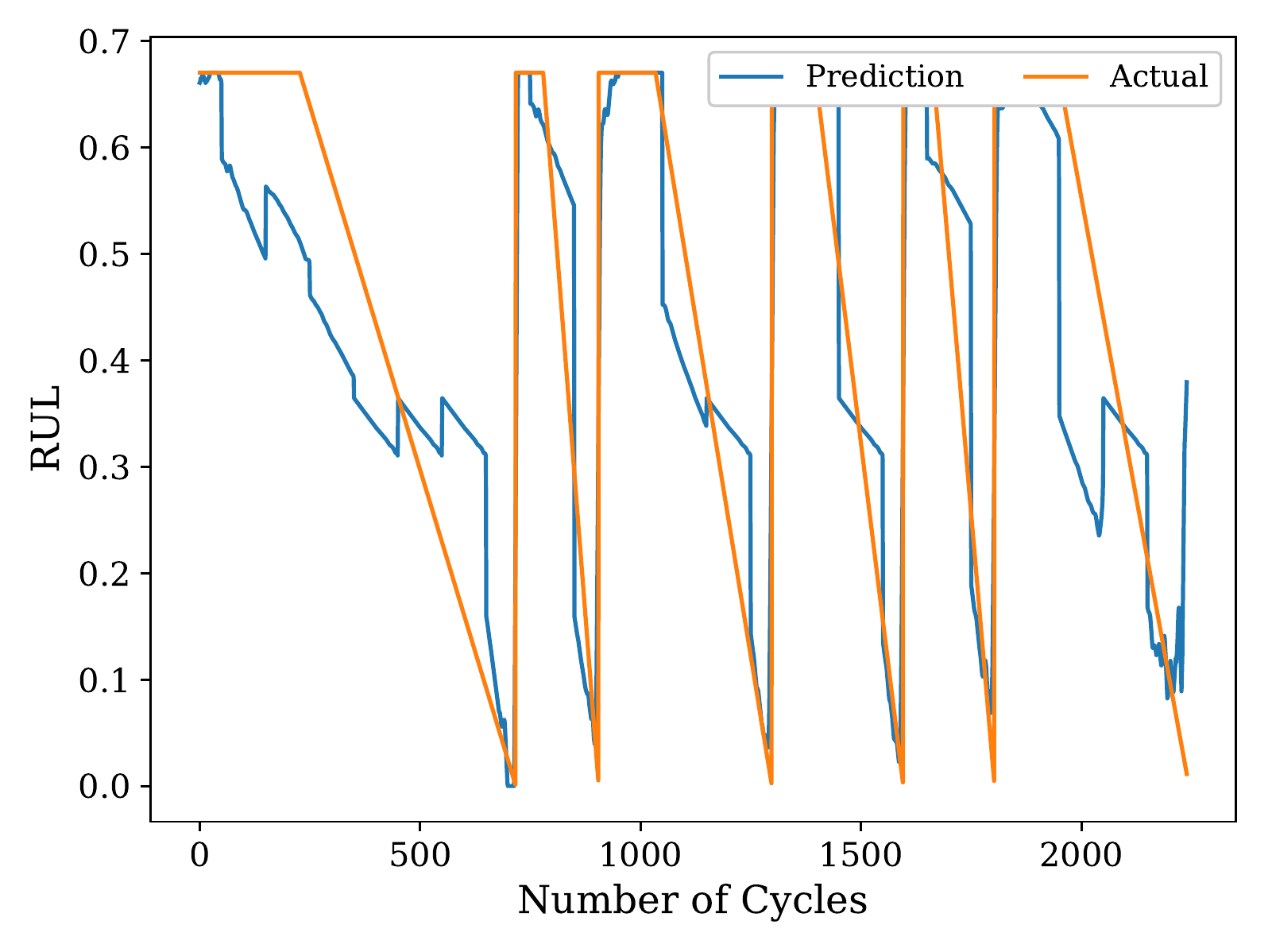}
	\end{minipage}
	}
	\subfigure[]
	{
	\begin{minipage}[t]{0.3\linewidth}
	\centering
	\includegraphics[width=5.5cm]{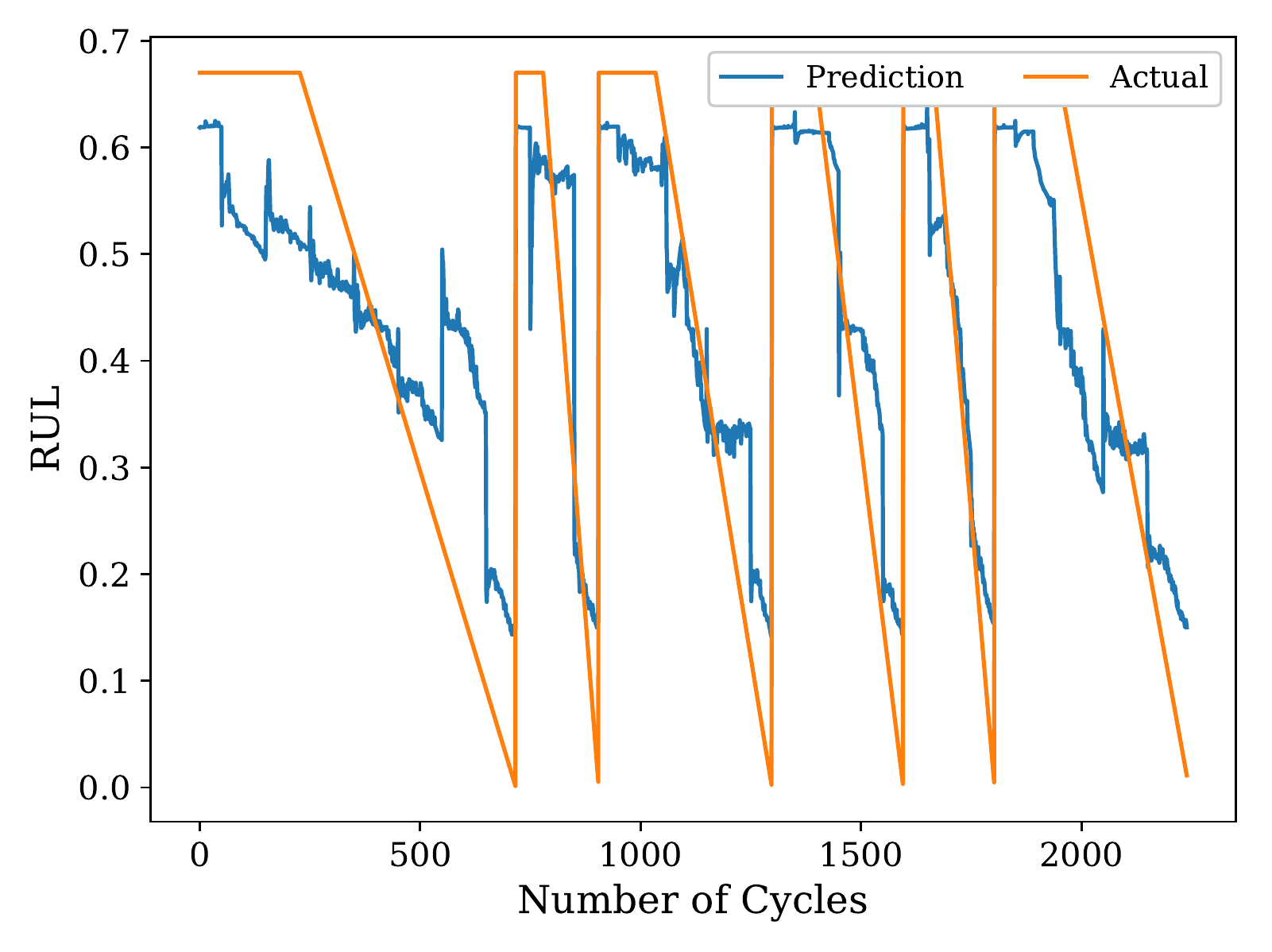}
	\end{minipage}
	}
	\caption{{RUL estimation comparison for degradation Pattern 1 with (a) LSTM, (b) CNN, and (c) RULNet .}}
	\label{FIGTII11}
\end{figure*}

\begin{table}[!ht]
\centering
\caption{Performance comparison for the proposed RULNet with three features using the index $RMSE$.}
\scriptsize
\renewcommand{\arraystretch}{1.5}
\begin{tabular}{c|c|c}
\hline
\hline
{\textbf{Feature Engineering}} & {\textbf{Pattern 1}} & {\textbf{Pattern 2}} \\
\hline
Features with health indicators  & {22.09} & \textbf{18.50} \\
\hline		
Features with SOM & {21.86} & NA  \\
\hline
Features with curve fitting & \textbf{20.61} & {20.08} \\
\hline
\hline
\end{tabular}
\label{tb:perf_table2}
\end{table}

It is easy to understand that the comparisons of different RUL estimation models are conducted on run-to-failure time-series. According to the clustering results of degradation patterns given in Section IV.A, there are two degradation patterns, and six RULNet models are trained accordingly, using the time window as 20 cycles. Therefore, two RULNet models per each feature methodology are trained and tested accordingly. For the training phase, sequence length is 100, and batch size is 32. Besides, Adam optimizer with a learning rate of 1e-4 is occupied. We trained the models over 5000 epochs, and the model with the lowest validation loss is used for performance evaluation.

To compare the performance of RULNet, another two neural network models, namely LSTM and CNN, are trained. The configuration of LSTM is the same as the sub-network starting from Layer 9 to Layer 15 in Table \ref{tb:system_model}. CNN models are similar to sub-network that consist of Layer 1 to Layer 8 followed by Layers 13, 14, and 15 in Table \ref{tb:system_model}.

\subsection{Performance Comparison}
With classified patterns and defined network configurations, the proposed RULNet is comprehensively compared with LSTM and CNN using three kinds of features. The index root mean square error ($RMSE$) is adopted to measure the accuracy of the RUL estimation result.

Table IV summarizes estimation error evaluated by the index $RMSE$ for all methods. We compare the performance of LSTM, CNN, and proposed RULNet under each feature engineering. Considering raw data are widely adopted in many recent literatures \cite{Lei2018Review}, we also present the estimation results using LSTM and CNN with raw data. It is observed that the proposed RULNet with any feature engineering outperforms LSTM and CNN with raw data for both Patterns 1 and 2. To evaluate the efficacy of different feature engineering ways, RULNet yields better performance using the features with health indicator and SOM, compared with LSTM and CNN. For the features with curve fitting, RULNet has similar results with CNN in Pattern 1, and they both show a better result than that of LSTM. Besides, RULNet achieves the best estimation accuracy in Pattern 2.

Performances of proposed RULNet with three feature engineerings are further compared, as given in Table \ref{tb:perf_table2}. For Pattern 1, features with curve fitting achieve the best performance since the lowest $RMSE$ surpass the performances with health indicators and SOM methodologies. Besides, in the case of enough training data, it infers that the more domain knowledge involved, the more accurate RUL estimation results. For Pattern 2, health indicators outperform features from curve fitting, which may be attributed to the following reason. The data amount of Pattern 2 is much less compared to Pattern 1, which may lead to inaccurate curve fitting parameters.

For clear understanding, the sample RUL estimation scenario of three models RULNet, LSTM, and CNN, are compared with the expected piece-wise RUL function. An example scenario of comparing RUL predictions from concatenated contact resistance in Pattern 1 data is shown in Fig. \ref{FIGTII11}. The estimated RUL graph from the RULNet model is more promising to follow the expected piece-wise RUL function compared to the prediction graphs from LSTM and CNN.

It is worth noting that the starting point is crucial to construct the piece-wise function. If expert knowledge is available, the starting point can be inferred correspondingly. Alternatively, multivariate statistical process analysis is capable of identifying the specific value of starting point by checking the variations \cite{WeibullFunction}, temporal dynamics \cite{2022ChagePoint}, etc.

\section{Conclusion}
In this paper, the remaining useful life (RUL) estimation of the reed relays was approached from the perspectives of degradation pattern analysis, and a series of features-based hybrid estimation models were proposed. The analysis of degradation behavior avoids simple recognition of a single pattern, contributing to designing a suitable model for a specific pattern. Then three typical feature engineerings are comprehensively compared, which are health indicators, features with self-organizing map, and features with curve fitting. The analysis helps to clarify a specific feature according to the data characteristics of employed data. Further, the unique abilities of long short-term network (LSTM) in capturing time dependencies and convolutional neural network (CNN) in extracting deep features are utilized to build a novel RULNet network, yielding the final RUL estimation results with the derived features. In a practical scenario, experiment studies demonstrated that the proposed method yields more accurate RUL estimation results in comparison with LSTM and CNN. Therefore, we can come to the conclusion that the usage of the RULNet model and the proposed hybrid approaches deliver a promising result for the problem of estimating RUL for reed relays according to the experimental result gained in this study.

In this article, we assume the starting point for online RUL prediction is known, which deserves more efforts to perform unsupervised health evaluation.

\ifCLASSOPTIONcaptionsoff
  \newpage
\fi


%








\end{document}